\def\year{2022}\relax
\documentclass[letterpaper]{article} 
\usepackage{aaai22}  
\usepackage{times}  
\usepackage{helvet}  
\usepackage{courier}  
\usepackage[hyphens]{url}  
\usepackage{graphicx} 
\usepackage{natbib}  
\usepackage{caption} 
\DeclareCaptionStyle{ruled}{labelfont=normalfont,labelsep=colon,strut=off} 
\usepackage{algorithm}
\usepackage{algorithmic}
\usepackage[flushleft]{threeparttable}
\usepackage{tablefootnote}
\usepackage{multirow}

\usepackage{stfloats}   
\usepackage[font=small, justification=centering]{caption}
\usepackage[font=small]{subcaption}
\usepackage[export]{adjustbox}
\usepackage{amsmath, amssymb}
\usepackage{amsfonts}
\usepackage{mathrsfs}
\DeclareMathOperator*{\argmax}{argmax}
\usepackage{algorithm}
\usepackage{algorithmic}
\usepackage{hyperref}

\usepackage{newfloat}
\usepackage{listings}
\floatstyle{ruled}
\newfloat{listing}{tb}{lst}{}
\floatname{listing}{Listing}
\title{Event-Aware Multimodal Mobility Nowcasting}
\author{
    Zhaonan Wang\textsuperscript{\rm 1,3$\dagger$}, 
    Renhe Jiang\textsuperscript{\rm 1,2\textsection},
    Hao Xue\textsuperscript{\rm 3},
    Flora D. Salim\textsuperscript{\rm 3},
    Xuan Song\textsuperscript{\rm 1},
    Ryosuke Shibasaki\textsuperscript{\rm 1}
}
\affiliations{
    \textsuperscript{\rm 1} Center for Spatial Information Science, University of Tokyo; \textsuperscript{\rm 2} Information Technology Center, University of Tokyo \\
    \textsuperscript{\rm 3} School of Computing Technologies, RMIT University \\
    \{znwang, jiangrh, songxuan, shiba\}@csis.u-tokyo.ac.jp, \{zhaonan.wang, hao.xue, flora.salim\}@rmit.edu.au
}

\usepackage{bibentry}

\begin{document}

\maketitle

\begingroup\renewcommand\thefootnote{$\dagger$}
\footnotetext{Work was done while the first author was virtually visiting \href{https://cruiseresearchgroup.github.io/}{the CRUISE research group} at RMIT University.}
\endgroup
\begingroup\renewcommand\thefootnote{\textsection}
\footnotetext{Corresponding author.}
\endgroup

\begin{abstract}
As a decisive part in the success of Mobility-as-a-Service (MaaS), spatio-temporal predictive modeling for crowd movements is a challenging task particularly considering scenarios where societal events drive mobility behavior deviated from the normality. While tremendous progress has been made to model high-level spatio-temporal regularities with deep learning, most, if not all of the existing methods are neither aware of the dynamic interactions among multiple transport modes nor adaptive to unprecedented volatility brought by potential societal events. In this paper, we are therefore motivated to improve the canonical spatio-temporal network (ST-Net) from two perspectives: (1) design a heterogeneous mobility information network (\textit{HMIN}) to explicitly represent intermodality in multimodal mobility; (2) propose a memory-augmented dynamic filter generator (\textit{MDFG}) to generate sequence-specific parameters in an on-the-fly fashion for various scenarios. The enhanced \underline{e}vent-\underline{a}ware \underline{s}patio-\underline{t}emporal \underline{net}work, namely \textbf{EAST-Net}, is evaluated on several real-world datasets with a wide variety and coverage of societal events. Both quantitative and qualitative experimental results verify the superiority of our approach compared with the state-of-the-art baselines. Code and data are published on \url{https://github.com/underdoc-wang/EAST-Net}.
\end{abstract}

\section{Introduction}
\begin{figure}[t]
	\centering
    \includegraphics[width=1.05\linewidth, center]{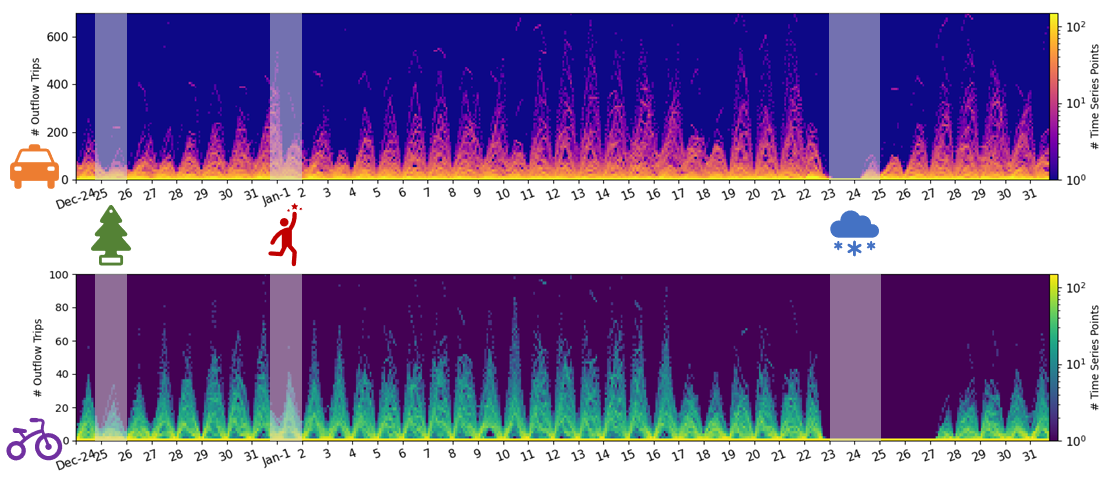}
	\caption{Time Series Histograms of Citywide \textit{Taxi} and \textit{Share Bike} Demands in Washington DC from 24 Dec. 2015 to 31 Jan. 2016, during which Christmas, New Year, and a Historic Blizzard ``\textit{Jonas}" Took Place}
	\vspace{-10pt}
	\label{fig:teaser}
\end{figure}

\noindent Mobility-as-a-Service (MaaS), as an emerging paradigm of transport service, seamlessly integrates multimodal mobility services (\textit{e.g.} public transport, ride-hailing, bike-sharing), which streamlines trip planning, ticketing (for users), operating optimization, emergency response (for providers), and traffic management (for city managers). For a smooth operation of MaaS, spatio-temporal predictive modeling for multimodal transport of crowds is indispensable. However, the existing methods either implicitly handle the interaction between supply and demand of different modes or assume it to be time-invariant \cite{ye2019co}. This task is even more challenging in scenarios where societal events (\textit{e.g.} holiday, severe weather, epidemic) take place and deviate collective mobility significantly from the normality (\textit{e.g.} daily, weekday routines). Moreover, as illustrated in Figure \ref{fig:teaser}, the impacts of different events differ, \textit{e.g.} taxi demand rockets on New Year's eve but vanishes at Christmas and during the blizzard, and the volatility brought to each transport mode varies, \textit{e.g.} recovery of share bike demand takes longer than the one of taxi after the blizzard.

While tremendous progress has been made in spatio-temporal modeling thanks to deep learning \cite{shi2015convolutional, zhang2017deep, li2018dcrnn_traffic, wu2019graph, zheng2020gman}, most, if not all, of them advance by exploiting high-level spatio-temporal regularities. The volatility brought by societal events, on the other hand, is by far downplayed and usually handled by simple rectifications, such as incorporating temporal covariates (\textit{e.g.} time-of-day, day-of-week, whether-holiday) as auxiliary input \cite{yao2018deep, zonoozi2018periodic}, adding a memory bank to reuse similar patterns in history \cite{yao2019learning, tang2020joint}. These manipulations to a certain degree bring time and holiday awareness, but they mainly help with the periodic and precedent parts and would still fail under more extreme scenarios like unprecedented events (\textit{e.g.} historic blizzard, COVID-19 pandemic). There is another line of research \cite{fan2015citymomentum, jiang2018deepurbanmomentum, jiang2019deepurbanevent} attempting to capture anomalous mobility tendency under events in an online fashion based on low-order Markov assumption and fine-grained time slot setting. These practices are arguably circumventing the inherent difficulty of the task instead of truly tackling it.

In this paper, we tackle the identified twofold unawareness of the existing spatio-temporal networks, namely intermodality-unaware and event-unaware, correspondingly via: (1) explicitly representing the dynamic interactions among multiple mobility modes; (2) intrinsically enhancing event-awareness and adaptivity of predictive models for various scenarios, including unprecedented events. Specifically, we design a heterogeneous information network to build the intermodal interactions into the widely adopted spatio-temporal modeling strategy; then leverage techniques of memory and dynamic filter networks that encourage the model to learn to distinguish and generalize to diverse scenarios. Based on the above two motivations, we propose an \underline{e}vent-\underline{a}ware \underline{s}patio-\underline{t}emporal \underline{net}work (\textbf{EAST-Net}). Our contributions are summarized as follows:
\begin{itemize}
    \item We design a new heterogeneous mobility information network (HMIN) to explicitly represent intermodal interactions (or intermodality) for spatio-temporal multimodal mobility modeling.
    \item We propose a novel memory-augmented dynamic filter generator (MDFG) to produce sequence-specific parameters on-the-fly, intrinsically improving event-awareness and adaptivity of spatio-temporal networks.
    \item We conduct a series of experiments on four real-world event-mobility datasets, and the results validate the superiority of EAST-Net quantitatively and qualitatively.
\end{itemize}

\section{Related Work}
Here we briefly review two lines of research: the first line is on modeling of event-related human mobility, while the other involves techniques for spatio-temporal forecasting and dynamic filter generation. (1) The former can be broken down into two branches, namely event-oriented and event-driven modeling. On the one hand, human mobility data (\textit{e.g.} GPS \cite{konishi2016cityprophet}, origin-destination records \cite{zhang2018detecting, zhang2019decomposition}, trip survey \cite{wang2021forecasting}) are commonly used as the underlying clues to infer both local and citywide anomalous events. On the other hand, mobility behavior is affected by these societal events conversely \cite{song2014prediction, fan2015citymomentum, jiang2019deepurbanevent, xie2020deep} and thereby deviates from normal patterns. (2) By effectively capturing complex non-linear dependency in the space and time, deep learning-based predictive models as a group outperforms classical statistical and matrix/tensor-based methods on both individual (\textit{e.g.} call detail records \cite{feng2018deepmove}, GPS trajectory \cite{fan2019decentralized}, Point-of-Interest visits \cite{xue2021mobtcast}) and collective mobility (\textit{e.g.} crowd volumes \cite{zhang2017deep}, demand of multimodal transport modes \cite{ye2019co}, origin-destination trips \cite{wang2019origin, jiang2021countrywide}) modeling. Besides, mainly studied for tasks like video prediction \cite{jia2016dynamic} and image classification \cite{yang2019condconv, zhou2021decoupled}, dynamic filter generation broadly shares a similar idea as model-based meta-learning (\textit{e.g.} memory-augmented neural networks \cite{santoro2016meta}, meta networks \cite{munkhdalai2017meta}) or hypernetworks \cite{ha2016hypernetworks}, by conditioning parameters of a target module on another network. Moreover, there have been some recent studies furthering the idea onto continual learning \cite{Oswald2020Continual}; meta knowledge-based parameterization \cite{pan2019urban} and spatial distinct filter generation \cite{cirstea2021enhancenet} for spatio-temporal forecasting.

\section{Preliminaries}
In this section, we firstly formulate \textit{multimodal mobility nowcasting} problem, then briefly revisit a standard solution, namely spatio-temporal network (ST-Net), for this task.

\subsection{Problem Definition}
Given a specified spatio-temporal granularity, the time and space can be discretized into a set of equal-length time slots and regions (not necessarily equal-area), respectively, denoted by $\mathscr{T} = \{\tau_t | t \in (1,\cdot\cdot\cdot, T)\}$ and $\mathscr{R} = \{\eta_n | n \in (1,\cdot\cdot\cdot, N) \}$. Considering there are in total $M$ modes of mobility, we can build a \textit{multimodal mobility tensor} $\underline{\mathcal{M}} \in \mathbb{R}^{T \times N \times C}$, where $C = 2 \cdot M$ if modeling the supply and demand of multiple transport modes; $C = M$ if modeling the visit volume of multiple travel purposes. Accordingly, \textit{multimodal mobility nowcasting} problem can be formulated as follows:

Given $\alpha$-step consecutive observations in $\underline{\mathcal{M}}$, denoted by $(\textbf{X}_{t-\alpha+1},\cdot\cdot\cdot,\textbf{X}_t)$, where $\textbf{X} \in \mathbb{R}^{N \times C}$, return the immediate expectations for the next $\beta$-step, \textit{i.e.} $(\hat{\textbf{X}}_{t+1},\cdot\cdot\cdot,\hat{\textbf{X}}_{t+\beta})$. Note that auxiliary temporal covariates can be available from time slot $\tau_{t-\alpha+1}$ to $\tau_{t+\beta}$, denoted by $\textbf{T}_{cov} \in \mathbb{R}^{(\alpha + \beta) \times v}$, where $v$ is total number of the covariates. Formally written as:
\begin{equation} \label{eq:problem}
    \begin{aligned}
	    \hat{\textbf{X}}_{t+1},\cdot\cdot\cdot,\hat{\textbf{X}}_{t+\beta} = &   \\ \mathop{\argmax}_{\textbf{X}_{t+1},\cdot\cdot\cdot,\textbf{X}_{t+\beta}} \textit{log P } & (\textbf{X}_{t+1},\cdot\cdot\cdot,\textbf{X}_{t+\beta} | \textbf{X}_{t-\alpha+1},\cdot\cdot\cdot,\textbf{X}_t; \textbf{T}_{cov})
    \end{aligned}
    \vspace{-13pt}
\end{equation}
\begin{figure}[h]
	\centering
	\begin{subfigure}{0.28\linewidth}
		\includegraphics[width=1.01\linewidth, center]{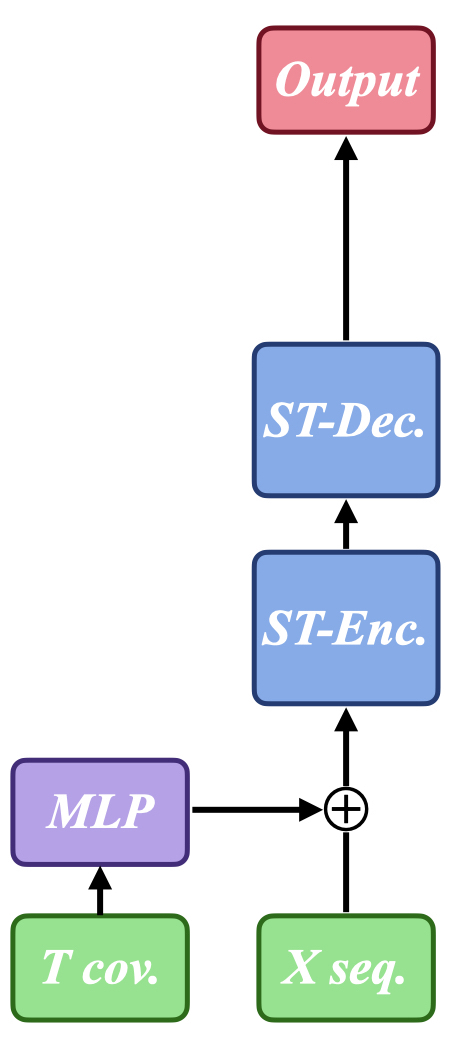}
		\caption{ST-Net$+\textbf{T}_{cov}$}
		\label{fig:event-model-1}
	\end{subfigure}
	\hspace{0.001cm}
	\begin{subfigure}{0.28\linewidth}
		\includegraphics[width=1.02\linewidth, center]{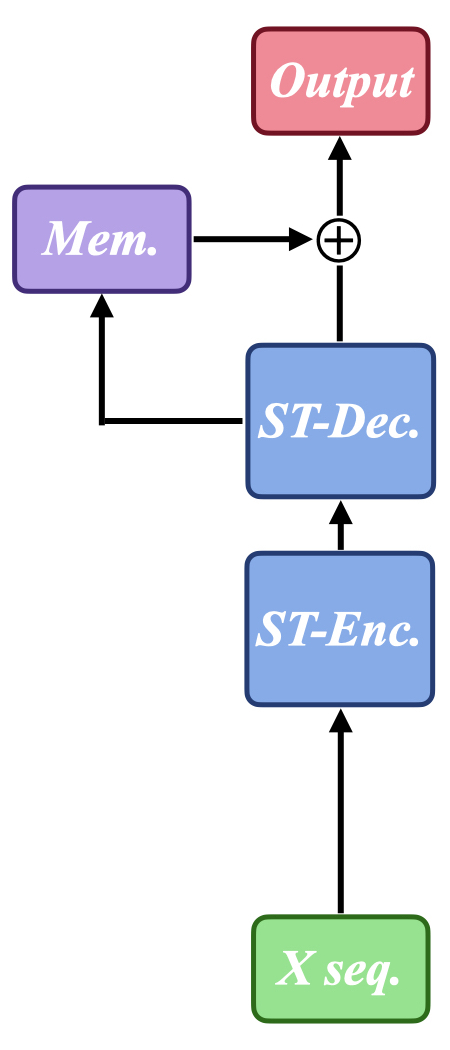}
		\caption{ST-Net$+$\textit{Mem.}}
		\label{fig:event-model-2}
	\end{subfigure}
	\hspace{0.001cm}
	\begin{subfigure}{0.38\linewidth}
		\includegraphics[width=1\linewidth, center]{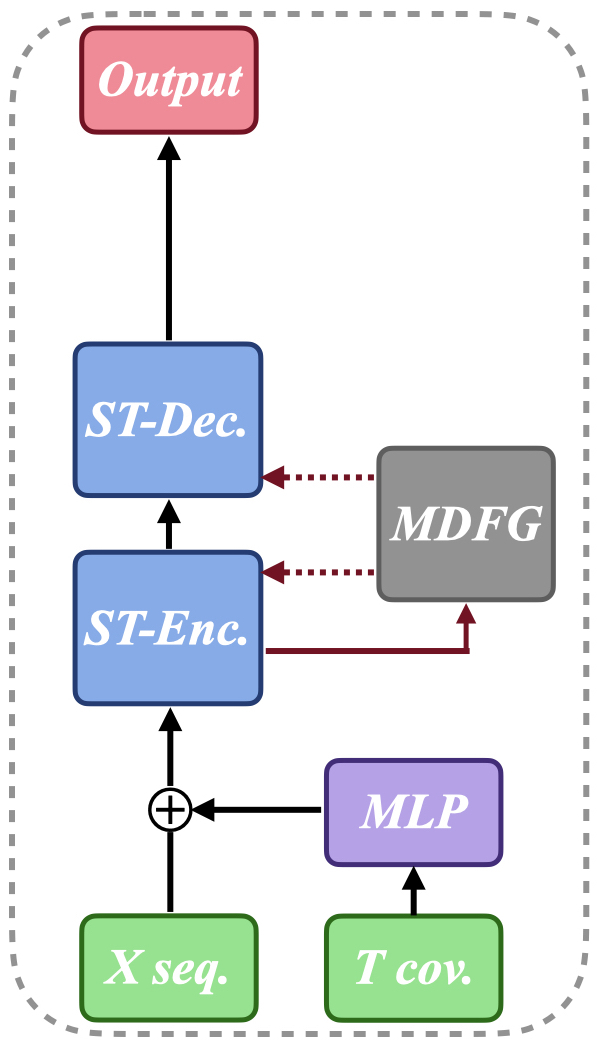}
		\caption{EAST-Net}
		\label{fig:event-model-3}
	\end{subfigure}
	\caption{Comparison of Abstract Structure between the Existing ST-Net with Rectifications (a) (b) and Proposed EAST-Net (c)}
	\vspace{-12pt}
	\label{fig:event-model}
\end{figure}

\subsection{Spatio-Temporal Network}
To solve the above problem, recent studies commonly exploit high-level spatio-temporal dependency in observations \cite{zhang2017deep, li2018dcrnn_traffic, ye2019co, wu2020connecting}. Particularly, convolutional and recurrent neural networks (\textit{e.g.} CNN, GCN, TCN, RNN) are two typical submodules utilized to handle the underlying dependencies over the space $\mathscr{R}$ and time$\mathscr{T}$, respectively. This class of models arguably share a similar spatio-temporal view, which prioritizes the first and second dimensions in $(\textbf{X}_{t-\alpha+1},\cdot\cdot\cdot,\textbf{X}_t) \in \mathbb{R}^{\alpha \times N \times C}$. We term this modeling strategy \textit{Spatio-Temporal Graph} (STG), as demonstrated in Figure \ref{fig:stg-hmin}, in which the third dimension of the observations is treated as features evolving on STG. We further term models built on top of STG \textit{Spatio-Temporal Network} (ST-Net). Without loss of generality, we combine GCN and RNN to denote a ST-Net, which handles spatial dependency by a graph and temporal dependency in a recurrent form:
\begin{equation} \label{eq:gcn}
	\textbf{H} = \sigma (\textbf{X} \star_\mathcal{G} \Theta) = \sigma (\mathop{\sum}_{k=0}^K \mathcal{\tilde P}^k \textbf{X} \textbf{W}_k)
\end{equation}
\begin{equation} \label{eq:gcru}
	\begin{cases}
	\begin{aligned}
		\textbf{u}_t & = sigmoid([\textbf{X}_t^{(l)}, \textit{ } \textbf{H}_{t-1}^{(l)}] \star_\mathcal{G} \Theta_{\textbf{u}} + b_{\textbf{u}})      \\
		\textbf{r}_t & = sigmoid([\textbf{X}_t^{(l)}, \textit{ } \textbf{H}_{t-1}^{(l)}] \star_\mathcal{G} \Theta_{\textbf{r}} + b_{\textbf{r}})      \\
		\textbf{C}_t & = tanh([\textbf{X}_t^{(l)}, \textit{ } (\textbf{r}_t \odot \textbf{H}_{t-1}^{(l)})] \star_\mathcal{G} \Theta_{\textbf{C}} + b_{\textbf{C}})      \\
		\textbf{H}_t^{(l)} & = \textbf{u}_t \odot \textbf{H}_{t-1}^{(l)} + (1 - \textbf{u}_t) \odot \textbf{C}_t
	\end{aligned}
	\end{cases}
\end{equation}
Equation \eqref{eq:gcn} defines the basic graph convolution operation $\star_\mathcal{G}$, which takes input $\textbf{X} \in \mathbb{R}^{N \times p}$ and returns $\textbf{H} \in \mathbb{R}^{N \times q}$ given a graph topology matrix $\mathcal{P} \in \mathbb{R}^{N \times N}$ ($\mathcal{\tilde P}$ is its normalized form), approximation order $K$, and trainable parameters $\Theta \in \mathbb{R}^{(K+1) \times p \times q}$. Equation \eqref{eq:gcru} defines an extended version of GRU (a form of RNN), namely GCRU, with matrix multiplications replaced by graph convolutions (Equation \eqref{eq:gcn}). It is noteworthy that DCGRU \cite{li2018dcrnn_traffic} can be seen as a special form of GCRU by restricting $\mathcal{\tilde P}$ to be random walk normalized transition matrix and performing bidimensional graph diffusion. Then, stacking multiple layers (denoted by $l$) of GCRU forms encoder and decoder of a ST-Net, abbreviated as \textit{ST-Enc/ST-Dec} in Figure \ref{fig:event-model}.

Besides, as illustrated in Figure \ref{fig:event-model-1}, temporal covariates can be used as auxiliary input \cite{yao2018deep, zonoozi2018periodic} to equip ST-Net with time and holiday awareness. In this case, $\textbf{X}_t^{(0)} = [\textbf{X}_t, \textbf{T}'_t]$, where $[,]$ denotes a concatenation operation and $\textbf{T}'_t$ is the linear projected representation of $\textbf{T}_{cov}$ at $\tau_t$. Another rectification for ST-Net (demonstrated in Figure \ref{fig:event-model-2}) attaches an external memory bank \cite{yao2019learning, tang2020joint} to the decoder such that some typical spatio-temporal patterns can be stored for reuse. This memory is implemented by a parameter matrix $\textbf{M} \in \mathbb{R}^{m \times D}$, where $m$ and $d$ denote the total number of memory records and dimension of each one. Before making the final output, decoder makes a query to $\textbf{M}$ for find similar representations, which is implemented by attention mechanism \cite{bahdanau2014neural, vaswani2017attention}. Formally,
\begin{equation} \label{eq:st-mem}
	\begin{cases}
	\begin{aligned}
		\text{Q}_t & = \overline{\textbf{H}}_t^{(l)} \textbf{W}_Q + b_Q     \\
		\phi_j & = \frac{e^{Q_t * \textbf{M}[j]}}{\sum_{j=1}^m e^{Q_t * \textbf{M}[j]}}     \\
		\tilde{\textbf{V}} & = (\sum_{j=1}^m \phi_j \cdot \textbf{M}[j]) \textbf{W}_V + b_V
	\end{aligned}
	\end{cases}
\end{equation}
where $\text{Q}_t \in \mathbb{R}^D$ denotes the query vector projected from flattened $\textbf{H}_t^{(l)}$; $\phi_j$ is the attention score corresponding to $j$-th memory record. The obtained vector $\tilde{\textbf{V}}$ can be reshaped back and concatenated with $\textbf{H}_t^{(l)}$ for output, $\textbf{H}_t^{(out)} = [\textbf{H}_t^{(l)}, \textbf{V}]$.

\section{Methodology}
In this section, we elaborate the motivations and techniques for improving ST-Net, and present Event-Aware Spatio-Temporal Network (EAST-Net) as a more adaptive framework for multimodal mobility nowcasting.

\subsection{Heterogeneous Mobility Information Network}
\begin{figure}[h]
	\centering
    \includegraphics[width=1.0\linewidth, center]{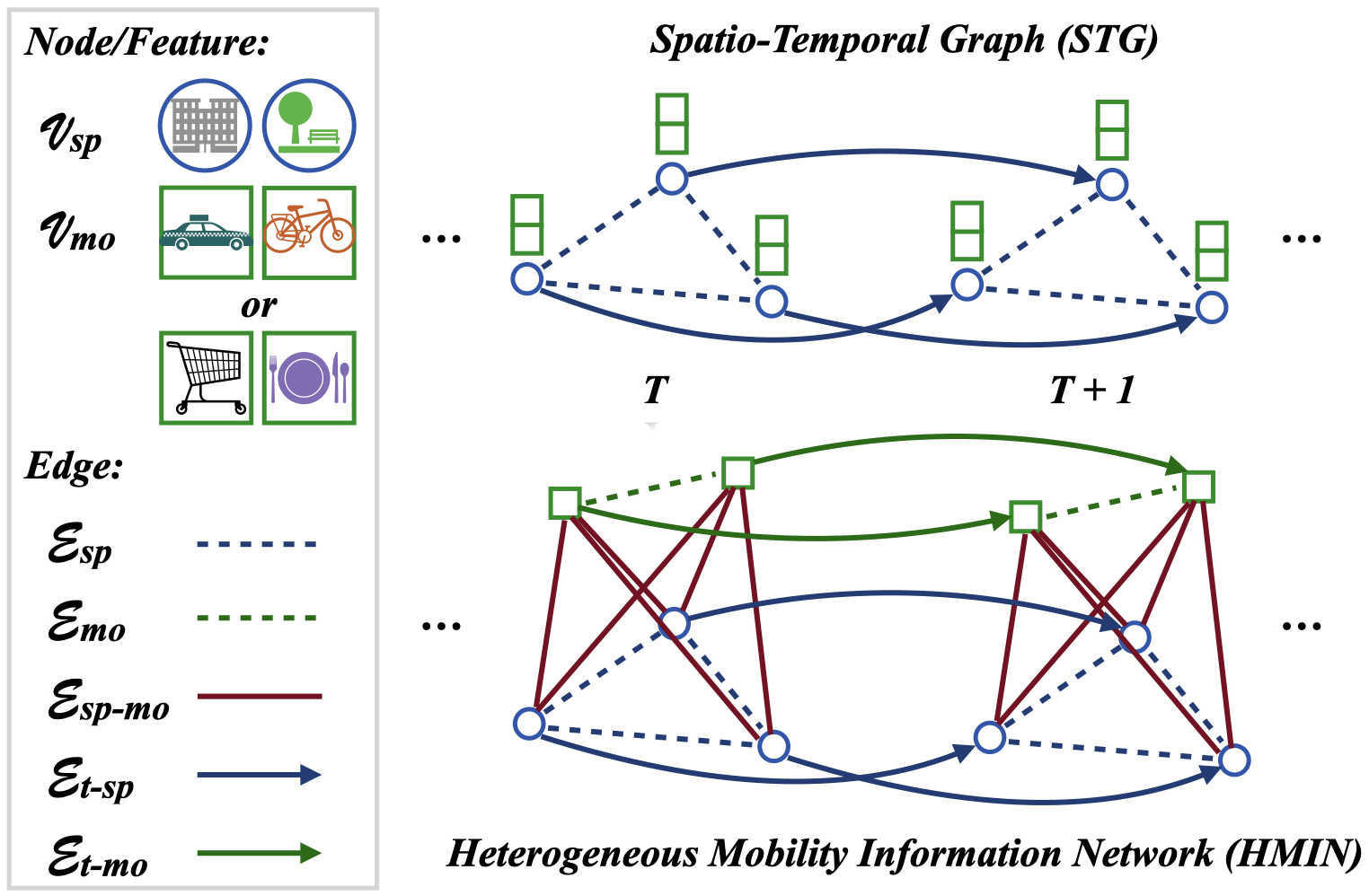}
	\caption{Comparison between Spatio-Temporal Graph (STG) and Heterogeneous Mobility Information Network (HMIN) for Multimodal Mobility Modeling}
	\label{fig:stg-hmin}
\end{figure}

As presented in Figure \ref{fig:stg-hmin}, STG, the fundamental of ST-Net, prioritizes spatio-temporal modeling while restricting all features (\textit{i.e.} mobility modes) to evolve together on the fixed STG. We argue that this spatio-temporal view restricts the modeling of dynamic interactions among different modes of mobility, which is in fact the operating mechanism of MaaS. As demonstrated in Figure \ref{fig:teaser}, a societal event may impact different transport modes variously, which confirms the necessity for intermodality modeling. Thus, we are motivated to design a new underlying structure, \textit{i.e. Heterogeneous Mobility Information Network} (HMIN), to jointly represent intermodal interaction and spatio-temporal dependency.

As illustrated in Figure \ref{fig:stg-hmin}, HMIN is defined as $\mathcal{G} = (\mathcal{V}_{sp} \cup \mathcal{V}_{mo}, \mathcal{E}_{sp} \cup \mathcal{E}_{mo} \cup \mathcal{E}_{sp\text{-}mo} \cup \mathcal{E}_{t\text{-}sp} \cup \mathcal{E}_{t\text{-}mo})$, where $\mathcal{V}_{sp} = \{\eta_1,\cdot\cdot\cdot,\eta_N\}$ and $\mathcal{V}_{mo} = \{\mu_1,\cdot\cdot\cdot,\mu_M\}$ denote node set of regions and mobility modes, respectively; $\mathcal{E}_{sp}$, $\mathcal{E}_{mo}$, $\mathcal{E}_{sp\text{-}mo}$, $\mathcal{E}_{t\text{-}sp}$, $\mathcal{E}_{t\text{-}mo}$ denote five edge sets for the relations in region-to-region, mode-to-mode, region-to-mode, time-to-region, time-to-mode. By this definition, the intermodal relationship and its dynamicity can be represented by $\mathcal{E}_{mo}$ and $\mathcal{E}_{t\text{-}mo}$; and the task of \textit{multimodal mobility nowcasting} is reformulated as a \textit{link prediction} task for edge set $\mathcal{E}_{sp\text{-}mo}$ ($|\mathcal{E}_{sp\text{-}mo}| = N \cdot C$) from $\tau_{t+1}$ to $\tau_{t+\beta}$.

Here we propose a simple yet generic framework to encode-decode HMIN by applying handy GCRU in a similar fashion to ST-Net. Denoting the framework of \textit{ST-Enc/ST-Dec} (GCRU in multi-layer) as $\textbf{H}_{t+1},\cdot\cdot\cdot,\textbf{H}_{t+\beta} = \text{GCRU}_{\text{Enc-Dec}} (\textbf{X}_{t-\alpha+1},\cdot\cdot\cdot,\textbf{X}_t)$, processing of HMIN can be decomposed into two views (\textit{i.e.} spatial and intermodal) followed by a stepwise fusion layer, formally denoted by:
\begin{equation} \label{eq:hminet}
	\begin{cases}
	\begin{aligned}
		\textbf{H}_{t+1}^{(sp)},\cdot\cdot\cdot,\textbf{H}_{t+\beta}^{(sp)} & = \text{GCRU}_{\text{Enc-Dec}}^{(sp)} (\textbf{X}_{t-\alpha+1},\cdot\cdot\cdot,\textbf{X}_t)      \\
		\textbf{H}_{t+1}^{(mo)},\cdot\cdot\cdot,\textbf{H}_{t+\beta}^{(mo)} & = \text{GCRU}_{\text{Enc-Dec}}^{(mo)} (\textbf{X}_{t-\alpha+1}^{\top},\cdot\cdot\cdot,\textbf{X}_t^{\top})      \\
        \hat{\textbf{X}}_{t+\varepsilon} & = \sigma (\textbf{H}_{t+\varepsilon}^{(sp)} \textbf{W}_{out} \textbf{H}_{t+\varepsilon}^{(mo) \top})   \\
	\end{aligned}
	\end{cases}
\end{equation}
where $\varepsilon \in (1, \cdot\cdot\cdot, \beta)$ denotes the step index within horizon $\beta$; $\textbf{H}_{t+\varepsilon}^{(sp)} \in \mathbb{R}^{N \times q}$ and $\textbf{H}_{t+\varepsilon}^{(mo)} \in \mathbb{R}^{C \times q}$ denote spatial and modal embeddings on $\mathcal{V}_{sp}$ and $\mathcal{V}_{mo}$ at time slot $\tau_{t+\varepsilon}$, respectively; $\textbf{W}_{out} \in \mathbb{R}^{q \times q}$ denotes a parameter matrix to fuse the node embeddings for link generation. For simplicity, we denote this framework by HMINet (Equation (\ref{eq:hminet})) and consider it to be a general case of ST-Net, which only takes the spatial view and let $\textbf{W}_{out} \in \mathbb{R}^{q \times C}$, $\textbf{H}_{t+\varepsilon}^{(mo)} = \textbf{I}_C$. Essentially, edge sets $\mathcal{E}_{sp}$ and $\mathcal{E}_{mo}$, representing spatial and intermodal dependencies, are handled by graph convolution in each domain; unidirectional temporal edges $\mathcal{E}_{t\text{-}sp}$ and $\mathcal{E}_{t\text{-}mo}$ are encoded by the recurrent structure; HMINet learns the mapping from $\alpha$-step to $\beta$-step in edge set $\mathcal{E}_{sp\text{-}mo}$.
\begin{figure*}[t]
	\centering
    \includegraphics[width=0.9\linewidth, center]{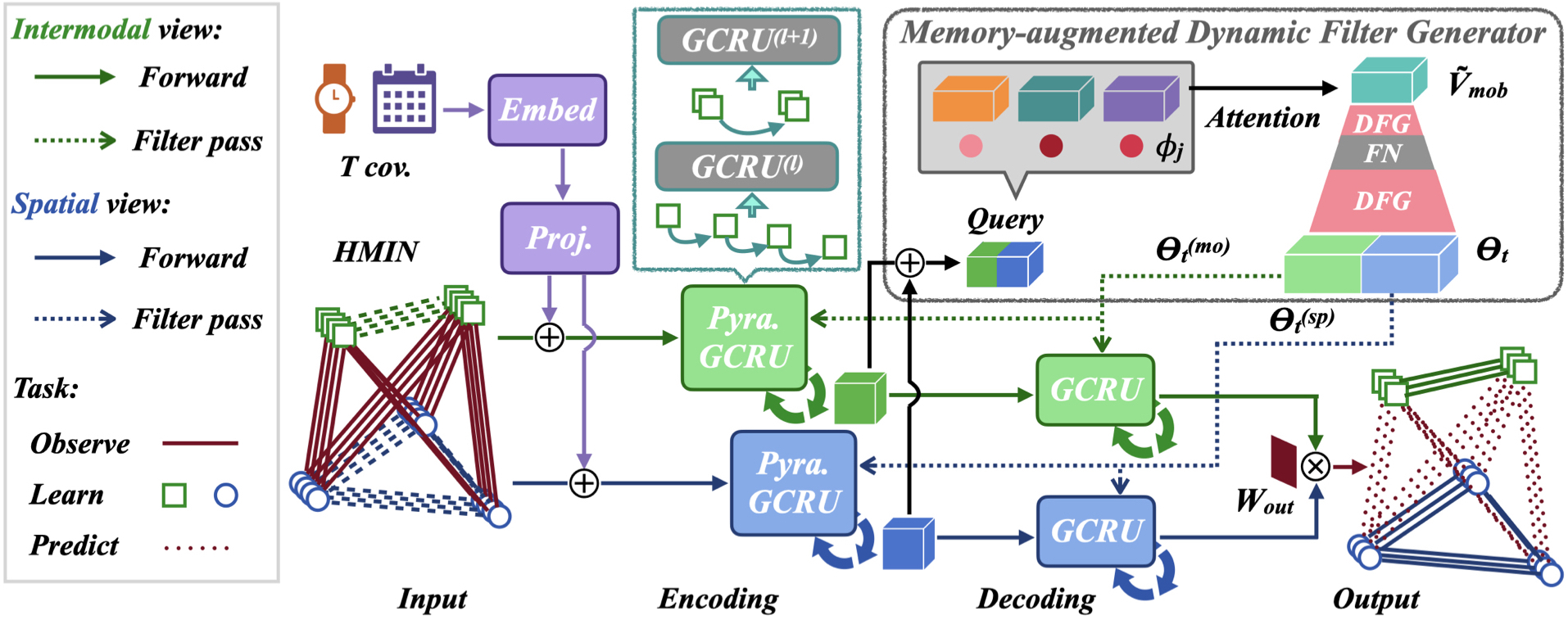}
	\caption{Framework of Proposed \textit{Event-Aware Spatio-Temporal Network}: \textbf{EAST-Net} (1) Takes a Sequence in Multimodal Mobility Tensor and Temporal Covariates as Inputs; (2) Encodes the Former Input in Heterogeneous Mobility Information Network (HMIN) with Two Branches of Pyramidal GCRU and Concatenate with the Latter; (3) Queries Memory-augmented Dynamic Filter Generator (MDFG) to Produce Sequence-specific Parameters as Graph Convolution Kernels; (4) Decodes with GCRU and Perform Link Prediction}
	\vspace{-5pt}
	\label{fig:model}
\end{figure*}

\subsection{Memory-augmented Dynamic Filter Generator}
Although enhancing ST-Net in an intermodality-aware way, HMINet introduces extra parameters by approximately same amount that ST-Net has. To control the model size and, more importantly, empower it to be aware of and adaptive to various scenarios, we propose a novel \textit{Memory-augmented Dynamic Filter Generator} (MDFG).

MDFG is motivated by a line of research on dynamic filter networks (DFN) \cite{jia2016dynamic, yang2019condconv, zhou2021decoupled}, which have been mainly studied on convolutional kernels for image and video-related tasks. The core idea behind DFN is instead of sharing a same trainable filter for all samples in a dataset, dynamically generating filters conditioned on an input sample, which by nature increases the flexibility and adaptivity of model. In light of DFN, we argue that the indistinguishability between normal and event scenarios roots in the way that a same set of parameters (\textit{e.g.} $\Theta' = [\Theta_{\textbf{u}}, \Theta_{\textbf{r}}, \Theta_{\textbf{C}}]$ in Equation (\ref{eq:gcru})) is shared for all observational sequences $(\textbf{X}_{t-\alpha+1},\cdot\cdot\cdot,\textbf{X}_t)$ by vanilla ST-Net. In other words, parameters in ST-Net are sequence-agnostic. We thereby utilize the idea of DFN and further put parameters conditioned on a plugin memory bank $\textbf{M}_{mob} \in \mathbb{R}^{m \times D}$ to encourage discovery of high-level mobility prototypes, which are representations incorporating spatial, temporal, and multimodal knowledge. To be specific, $\textbf{M}_{mob}$ takes concatenated node embeddings $[\textbf{H}_t^{(sp)}, \textbf{H}_t^{(mo)}] \in \mathbb{R}^{(N+C) \times q}$ as a query and returns a reconstructed prototype vector $\tilde{\textbf{V}}_{mob} \in \mathbb{R}^D$, which further passes through a DFN to produce momentary filters $[\Theta_t^{(sp)}, \Theta_t^{(mo)}]$ for $\text{GCRU}_{\text{Enc-Dec}}^{(sp)}$ and $\text{GCRU}_{\text{Enc-Dec}}^{(mo)}$. This interaction between HMINet and MDFG occurs in an on-the-fly manner, which generates sequence-specific parameters (denoted by $\Theta_t$). Formally,
\begin{equation} \label{eq:mdfg}
	\begin{cases}
	\begin{aligned}
		\text{Q}_t & = [\overline{\textbf{H}}_t^{(sp)}, \overline{\textbf{H}}_t^{(mo)}] \textbf{W}_Q + b_Q     \\
		\phi_j & = \frac{e^{Q_t * \textbf{M}_{mob}[j]}}{\sum_{j=1}^m e^{Q_t * \textbf{M}_{mob}[j]}}     \\
		\tilde{\textbf{V}}_{mob} & = \sum_{j=1}^m \phi_j \cdot \textbf{M}_{mob}[j]      \\
		\Theta_t & = [\Theta_t^{(sp)}, \Theta_t^{(mo)}] = \int(\varphi(\int(\tilde{\textbf{V}}_{mob})))
	\end{aligned}
	\end{cases}
\end{equation}
where $\int$ denotes a dynamic filter generation (DFG) layer, which can be implemented in various ways. Without loss of generality, we utilize a linear projection in this case. $\varphi$ denotes a filter normalization (FN) layer \cite{zhou2021decoupled}, used to normalize the generated parameters and avoid gradient vanishing and exploding.

\subsection{Event-Aware Spatio-Temporal Network}
Based on HMINet and MDFG, we further make three refinements to the framework of \textit{Event-Aware Spatio-Temporal Network} (EAST-Net), as illustrated in Figure \ref{fig:model}, which can be trained in an end-to-end fashion by minimizing a specified loss function using the standard backpropagation.
\begin{itemize}
    \item \textit{Temporal covariates} are fused stepwise for basic time and holiday awareness for both spatial and intermodal views, following the common practice \cite{yao2019learning}.
    \begin{equation} \label{eq:tcov_in}
	    \textbf{X}_{t+\varepsilon}^{(0)} = 
		\begin{cases}
			[\textbf{X}_{t+\varepsilon}, \textbf{T}'_{t+\varepsilon}] & \textit{, if }\varepsilon \in (1-\alpha, \cdot\cdot\cdot, 0)  \\
			[\hat{\textbf{X}}_{t+\varepsilon}, \textbf{T}'_{t+\varepsilon}] & \textit{, if }\varepsilon \in (1, \cdot\cdot\cdot, \beta)
		\end{cases}
    \end{equation} 
    Then, $\textbf{X}_{t+\varepsilon}^{(0)}$ is fed into HMINet (Equation (\ref{eq:hminet})) as input.
    \item \textit{Pyramidal structure} \cite{zonoozi2018periodic} is leveraged in GCRU encoders to help accelerate the training of HMINet and discover multi-level temporal pattern for mobility prototype extraction. In a case by a factor of $2$:
    \begin{equation}    \label{eq:pyra}
        \textbf{H}_t^{(l+1)} = [\textbf{H}_{2t}^{(l)}, \textbf{H}_{2t+1}^{(l)}]
    \end{equation}
    
    \item \textit{Adaptive edge sets} $\mathcal{E}_{sp}, \mathcal{E}_{mo}$ are learnt in HMINet without making any prior assumptions on either intermodal or spatial relationship \cite{li2018dcrnn_traffic, wu2019graph}. Essentially, a pair of parameterized node embeddings are initialized for both $\text{GCRU}_{\text{Enc-Dec}}^{(sp)}$ and $\text{GCRU}_{\text{Enc-Dec}}^{(mo)}$ to derive corresponding topology for graph convolutions:
    \begin{equation} \label{eq:graph_gen}
	\begin{cases}
	\begin{aligned}
		\tilde{\mathcal{E}}_{(sp)} & = softmax(relu(\textbf{E}_{(sp)} \textbf{F}_{(sp)}^{\top}))   \\
		\tilde{\mathcal{E}}_{(mo)} & = softmax(relu(\textbf{E}_{(mo)}  \textbf{F}_{(mo)}^{\top}))
	\end{aligned}
	\end{cases}
    \end{equation}
    where embeddings $\textbf{E}_{(sp)}, \textbf{F}_{(sp)} \in \mathbb{R}^{N \times \mu_{sp}}$ and $\textbf{E}_{(mo)},$ $\textbf{F}_{(mo)} \in \mathbb{R}^{C \times \mu_{mo}}$ are trained to learn the underlying region-to-region and mode-to-mode dependencies within node sets $\mathcal{V}_{sp}$ and $\mathcal{V}_{mo}$; the derived topology is normalized to $[0, 1]$ by softmax to simulate signal diffusion in each domain (replacing $\mathcal{\tilde P}$ in Equation (\ref{eq:gcn})).
\end{itemize}

\section{Experiments}
\begin{table*}[b]
    \footnotesize
    \begin{threeparttable}
    \centering
    \caption{Summary of Four Experimental Datasets}
    \label{tab:data}
    \addtolength{\tabcolsep}{-0.5ex}
        \begin{tabular}{c||cc|cc|cc|cc} 
            \hline
            \textbf{Dataset} & \multicolumn{2}{c|}{\textbf{JONAS-NYC}} & \multicolumn{2}{c|}{\textbf{JONAS-DC}} & \multicolumn{2}{c|}{\textbf{COVID-CHI}} & \multicolumn{2}{c}{\textbf{COVID-US}}  \\
            \hline
            Time Span & \multicolumn{2}{c|}{$2015/10/24 \sim 2016/1/31$} & \multicolumn{2}{c|}{$2015/10/24 \sim 2016/1/31$} & \multicolumn{2}{c|}{$2019/7/1 \sim 2020/12/31$} & \multicolumn{2}{c}{$2019/11/14 \sim 2020/5/31$}     \\
            Temporal & $100$ days & by $30$-minute & $100$ days & by $1$-hour & $550$ days & by $2$-hour & $200$ days & by $1$-hour     \\
            Spatial & $16 \times 8$ grid & in $0.5 \times 0.5$\textit{km} & $9 \times 12$ grid & in $0.5 \times 0.5$\textit{km} & $14 \times 8$ grid & in $1.5 \times 1.2$\textit{km} & \multicolumn{2}{c}{$50$ states $+$ DC}    \\
            \hline
            Mobility & \multicolumn{6}{c|}{\{Demand, Supply\} of Transport Mode in \{Taxi, Share Bike, Scooter*\}} & \multicolumn{2}{c}{Travel Purpose$^\star$}     \\
            \hline
            Event & \multicolumn{4}{c|}{Holidays $+$ \textbf{Blizzard Jonas} ($2016/1/22 \sim 24$)} & \multicolumn{4}{c}{Holidays $+$ \textbf{COVID Pandemic}$^\dagger$}        \\
        \hline
    \end{tabular}
\begin{tablenotes}
	\small
	\item * Scooter trip data is only available in \textbf{COVID-CHI} set.
	\item $^\star$ Travel purpose is measured by POI visitations of $10$ categories: \{grocery store, retailer, transportation, office, school, healthcare, entertainment, hotel, restaurant, service\}, according to the NAICS industry codes (\url{https://www.naics.com/search-naics-codes-by-industry/}).
	\item $^\dagger$ COVID-19 pandemic outbroke in late March 2020; \textbf{COVID-US} set depicts \textbf{the early stage} (first wave in April), and \textbf{COVID-CHI} set depicts \textbf{the progression} (first to third waves till end of 2020) of the pandemic.
\end{tablenotes}
\end{threeparttable}
\end{table*}
\subsection{Datasets}
To evaluate the proposed EAST-Net, we collect four real-world datasets with different spatio-temporal scales and coverage (presented in Table \ref{tab:data}), and represent \textit{multimodal mobility} with \textit{transport modes} on three city-level datasets (for New York City, Washington DC, Chicago), and with \textit{travel purpose} on the other country-level dataset (for the United States). Similarly to the previous studies \cite{zhang2017deep, ye2019co, jiang2019deepurbanevent}, trip records (\textit{e.g.} taxi, share bike) or POI visits are processed as in/outflow (supply/demand) or visit volume to be further aggregated onto a given spatio-temporal granularity.

Particularly, each dataset is designed to cover a set of holidays and a historic event with big social impact, \textit{i.e.} the winter storm Jonas or COVID-19 pandemic. Following the common practice \cite{zonoozi2018periodic, yao2018deep}, we encode temporal covariates of each time slot (\textit{i.e.} \textit{time-of-day}, \textit{day-of-week}, \textit{month-of-year}, \textit{whether-holiday}) in an one-hot manner as auxiliary sequence input.

\subsection{Settings}
We chronologically split each dataset for training, validation, testing with a ratio of $7:1:2$, such that the lengths of test sets are roughly last $20$ days for JONAS-\{NYC, DC\}, $110$ days for COVID-CHI, and $40$ days for COVID-US. Lengths of observational and nowcasting sequences are set to $\alpha = 8$ and $\beta = 8$, respectively; number of GCRU layers $L = 2$ with approximation order $K = 3$ and hidden dimension $q = 32$; embedding dimensions for $\textbf{T}_{cov}$ $v' = 2$, $\mu_{(sp)} = 20$ and $\mu_{(mo)} = 3$; mobility prototype memory $m = 8$ and $D = 16$. For model training, batch size $= 32$; learning rate $= 5 \times 10^{-4}$; maximum epoch $= 100$ with an early stopper with a patience of $10$; \textit{MAE} is chosen to be optimized using \textit{Adam}. We implement EAST-Net with \textit{PyTorch} and carry out experiments on a GPU server with \textit{NVIDIA GeForce GTX 1080 Ti} graphic cards. For evaluation, we adopt three commonly used metrics, namely Root Mean Square Error (\textit{RMSE}), Mean Absolute Error (\textit{MAE}) and Mean Absolute Percentage Error (\textit{MAPE}).

\subsection{Evaluations}
In this section, to understand the performance of our approach, we develop a group of research questions (RQ) and design a series of experiments correspondingly: (1) How does EAST-Net perform compared with the existing methods? (2) How does EAST-Net perform compared with its model variants? (3) How does EAST-Net behave in different scenarios of societal events?

\subsubsection{Quantitative Evaluation 1}
To quantitatively evaluate the overall prediction accuracy of EAST-Net on the multimodal mobility nowcasting problem, we implement eight baselines on mobility/traffic-related spatio-temporal prediction for comparison, including:
\begin{itemize}
    \item \textbf{Historical Average (HA):} Average values of same time slot in the training set for prediction.
    \item \textbf{Naive Forecast (NF):} Naively repeat the latest one observation for the next $\beta$ time slots. This practice is proven to be rather effective under events \cite{jiang2019deepurbanevent}.
    \item \textbf{Transformer$^\dagger$} \cite{vaswani2017attention}: The enhanced version \cite{li2019enhancing} with convolutional self-attention is implemented to capture local temporal pattern for time series forecasting.
    \item \textbf{CoST-Net} \cite{ye2019co}: A two-stage co-predictive model for multimodal transport demands. It models each mode individually with convolutional auto-encoder and uses a heterogeneous LSTM for collaborative modeling.
    \item \textbf{DCRNN} \cite{li2018dcrnn_traffic}: A special form of GCRU requiring a pre-defined transition matrix as auxiliary input to perform bidimensional graph diffusion.
    \item \textbf{Graph WaveNet (GW-Net)} \cite{wu2019graph}: A benchmark traffic forecasting model, in which parameterized graph input is firstly proposed. It utilizes a WaveNet-like structure for temporal modeling.
    \item \textbf{MTGNN} \cite{wu2020connecting}: A state-of-the-art model for multivariate time series modeling. It features an efficient unidirectional graph constructor and multi-kernel TCN.
    \item \textbf{StemGNN} \cite{cao2020spectral}: Another state-of-the-art method by modeling spatial and temporal dependencies jointly in the spectral domain for multivariate time series (MTS) forecasting.
\end{itemize}
\begin{table*}[t]
    \footnotesize
    \centering
    \caption{Performance Comparison of EAST-Net and Baselines in \textit{RMSE}, \textit{MAE}, \textit{MAPE} at JONAS-\{NYC, DC\}, COVID-\{CHI, US\}}
    \label{tab:baseline}
    \addtolength{\tabcolsep}{-0.5ex}
        \begin{tabular}{c||ccc|ccc|ccc|ccc} 
            \hline
            \multirow{2}{*}{\textbf{Model}} & \multicolumn{3}{c|}{\textbf{JONAS-NYC}} & \multicolumn{3}{c|}{\textbf{JONAS-DC}} & \multicolumn{3}{c|}{\textbf{COVID-CHI}} & \multicolumn{3}{c}{\textbf{COVID-US}}  \\
            \cline{2-13}
            & \textit{RMSE} & \textit{MAE} & \textit{MAPE} & \textit{RMSE} & \textit{MAE} & \textit{MAPE} & \textit{RMSE} & \textit{MAE} & \textit{MAPE} & \textit{RMSE} & \textit{MAE} & \textit{MAPE}  \\
            \hline
            HA & 48.953 & 39.221 & 75.33\% & 6.316 & 3.112 & \underline{38.86\%} & 37.156 & 9.938 & 190.48\% & 2822.12 & 1218.61 & 159.72\%    \\
            NF & 29.928 & 28.370 & 59.25\% & 7.754 & 3.594 & 68.95\% & 12.909 & 5.662 & 79.19\% & 2385.28 & 1258.12 & 185.08\%  \\
            Transformer$^\dagger$ & 35.050 & 23.428 & 47.08\% & 6.544 & 2.963 & 65.73\% & 12.671 & 5.106 & 80.50\% & 1767.71 & 862.82 & 180.13\%      \\
            CoST-Net & 33.721 & 22.485 & 41.03\% & 6.274 & 2.971 & 52.05\% & 15.259 & 6.881 & 83.74\% & - & - & -      \\
            DCRNN & 28.722 & \underline{18.718} & 38.99\% & 5.469 & 3.066 & 50.35\% & 10.566 & 6.483 & 51.23\% & 1194.38 & 722.34 & 155.92\%     \\
            GW-Net & \underline{28.584} & 19.367 & 36.96\% & \underline{5.091} & \underline{2.334} & 51.03\% & \textbf{8.365} & \underline{3.723} & \textbf{45.41\%} & \underline{1022.82} & \underline{490.97} & 77.62\%     \\
            MTGNN & 28.874 & 19.118 & \underline{36.39\%} & 5.161 & 2.691 & 47.95\% & 8.822 & 4.350 & 51.58\% & 1083.00 & 535.61 & \underline{75.86\%}    \\
            StemGNN & 30.711 & 21.489 & 40.81\% & 5.316 & 3.074 & 50.44\% & \underline{8.400} & 4.496 & \underline{50.27\%} & 1279.04 & 709.16 & 146.73\%     \\
            \hline
            \textbf{EAST-Net} & \textbf{23.632} & \textbf{15.790} & \textbf{33.33\%} & \textbf{4.103} & \textbf{2.004} & \textbf{35.03\%} & 9.381 & \textbf{3.380} & 61.50\% & \textbf{799.51} & \textbf{371.78} & \textbf{51.84\%}      \\
            -$\Delta$\% & -17.3\% & -15.6\% & -8.4\% & -19.4\% & -14.1\% & -9.9\% & - & -9.2\% & - & -21.8\% & -24.3\% & -31.7\%    \\
        \hline
    \end{tabular}
    \vspace{-10pt}
\end{table*}

We present the performance comparison of EAST-Net and baselines in Table \ref{tab:baseline}. It is noticeable that the error range on four datasets varies in magnitude: among three city-level sets, DC and CHI have relatively smaller transport volume than NYC; COVID-US is apparently the most tricky set which is state-level, of ten modes for travel purpose, and being tested at the very early stage (first wave) of the pandemic. Besides, acceptable results obtained by HA on JONAS-DC, NF on JONAS-NYC and COVID-CHI indicate a rather strong short-term temporal dependency in JONAS-NYC and COVID-CHI, and a daily periodicity in JONAS-DC. By treating the problem simply as time series, Transformer does not acquire satisfactory accuracy. Taking spatial locality into consideration, CoST-Net performs better than Transformer on JONAS-\{NYC, DC\}, but the pre-trained convolutional structure not only fails it on COVID-CHI but limits it from handling graph-based data like COVID-US. Then, among four graph-based models, GW-Net prevails in terms of most metrics on all datasets. Lastly, speaking of EAST-Net, we can observe a consistent and dramatic improvement throughout JONAS-\{NYC, DC\} and COVID-US, which undoubtedly confirms the efficacy of EAST-Net. The exception on COVID-CHI, we think, can be explained by: (1) A coarse time slot (2-hour) setting ``smoothes'' sudden changes, making the task easier for other models; (2) Along with the progression (first to third waves) of COVID pandemic, other models gradually learn the pandemic pattern as a new normality.

\subsubsection{Quantitative Evaluation 2}
To understand how EAST-Net improves from the canonical ST-Net, we implement ST-Net (in Equation \eqref{eq:gcru}) and its two rectified forms (in Figure \ref{fig:event-model-1} and \ref{fig:event-model-2}), as well as HMINet (in Equation \eqref{eq:hminet}) for comparison. As presented in Table \ref{tab:ablation}, within the ST-Net family, a regular memory bank improves ST-Net in most cases, but not as significantly as temporal covariates do. However, adding $\textbf{T}_{cov}$ deteriorates the performance on COVID-US, which is actually reasonable because the reinforced awareness of periodicity backfires especially at the early stage of a historic epidemic when human mobility began to deviate (because of quarantine measures). In comparison, adopting HMINet drops all metrics compared with regular ST-Net especially on COVID-US, which validates our motivation for explicit intermodality modeling. Besides, adopting HMIN on COVID-CHI seems not as helpful as on other datasets. This issue, we think, may be caused by including the scooter data, which is in fact a pilot program in Chicago and thus has some months without any data. Lastly, comparing HMINet and EAST-Net side by side, we can observe a consistent performance improvement, which verifies the effectiveness of MDFG in various scenarios.
\begin{table*}[t]
    \footnotesize
    \centering
    \caption{Performance Comparison of EAST-Net and Variants in \textit{RMSE}, \textit{MAE}, \textit{MAPE} at JONAS-\{NYC, DC\}, COVID-\{CHI, US\}}
    \label{tab:ablation}
    \addtolength{\tabcolsep}{-0.5ex}
        \begin{tabular}{c||ccc|ccc|ccc|ccc} 
            \hline
            \multirow{2}{*}{\textbf{Variant}} & \multicolumn{3}{c|}{\textbf{JONAS-NYC}} & \multicolumn{3}{c|}{\textbf{JONAS-DC}} & \multicolumn{3}{c|}{\textbf{COVID-CHI}} & \multicolumn{3}{c}{\textbf{COVID-US}}  \\
            \cline{2-13}
            & \textit{RMSE} & \textit{MAE} & \textit{MAPE} & \textit{RMSE} & \textit{MAE} & \textit{MAPE} & \textit{RMSE} & \textit{MAE} & \textit{MAPE} & \textit{RMSE} & \textit{MAE} & \textit{MAPE}  \\
            \hline
            ST-Net & 31.382 & 20.215 & 43.99\% & 5.437 & 2.366 & 55.96\% & 12.166 & 5.061 & 80.00\% & 1123.91 & \underline{519.07} & \underline{62.17\%}    \\
            ST-Net$+\textbf{T}_{cov}$ & \underline{25.353} & \underline{16.964} & \underline{33.97\%} & \underline{4.453} & \underline{2.042} & \underline{43.05\%} & \textbf{8.674} & \textbf{2.823} & \underline{59.00\%} & 1434.33 & 720.08 & 82.42\%      \\
            ST-Net$+$\textit{Mem.} & 30.725 & 20.158 & 40.41\% & 5.079 & 2.599 & 44.00\% & 9.921 & \underline{3.018} & \textbf{57.00\%} & \underline{1058.52} & 528.29 & 63.36\%     \\
            \hline
            HMINet & 28.713 & 18.205 & 37.96\% & 4.567 & 2.072 & 48.26\% & 11.437 & 4.475 & 78.53\% & 906.85 & 399.35 & \textbf{43.47\%}     \\
            \textbf{EAST-Net} & \textbf{23.632} & \textbf{15.790} & \textbf{33.33\%} & \textbf{4.103} & \textbf{2.004} & \textbf{35.03\%} & \underline{9.381} & 3.380 & 61.50\% & \textbf{799.51} & \textbf{371.78} & 51.84\%      \\
        \hline
    \end{tabular}
    \vspace{-8pt}
\end{table*}
\begin{figure}[t]
	\centering
	\begin{subfigure}{1\linewidth}
		\includegraphics[width=1\linewidth, center]{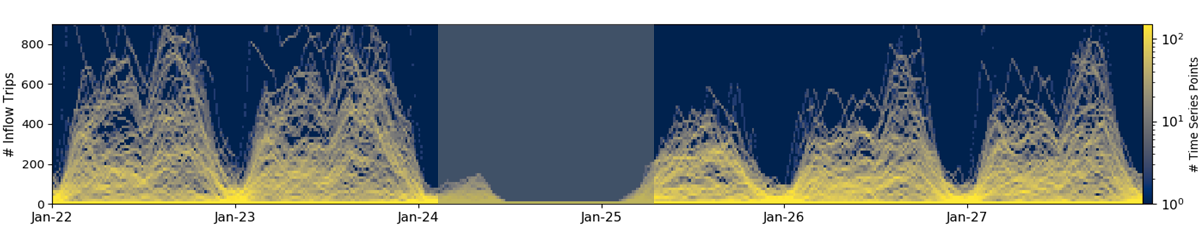}
		\vspace{-18pt}
		\caption{Ground Truth}
		\label{fig:case-nyc-true}
	\end{subfigure}
	\hspace{0.01cm}
	\begin{subfigure}{1\linewidth}
		\includegraphics[width=1\linewidth, center]{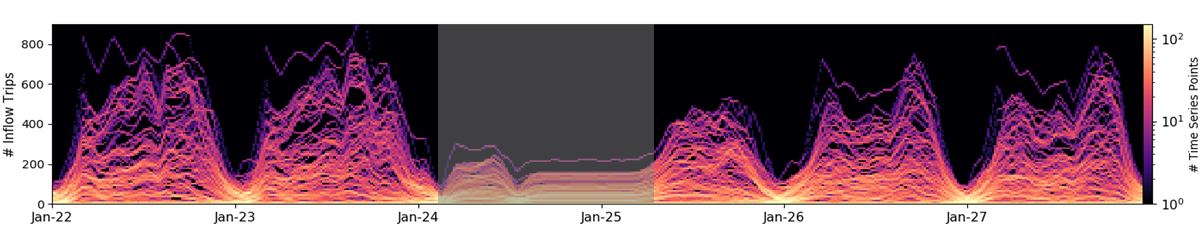}
		\vspace{-18pt}
		\caption{Prediction by GW-Net}
		\label{fig:case-nyc-sota}
	\end{subfigure}
	\hspace{0.01cm}
	\begin{subfigure}{1\linewidth}
		\includegraphics[width=1\linewidth, center]{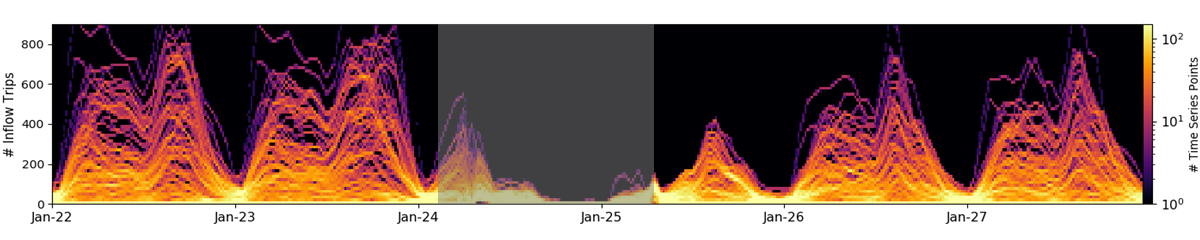}
		\vspace{-18pt}
		\caption{Prediction by EAST-Net}
		\label{fig:case-nyc-east}
	\end{subfigure}
	\vspace{-5pt}
	\caption{Time Series Histograms of the Ground Truth and (2-hour ahead) Prediction Results of Citywide Taxi Demand in New York City from 22 Jan. 2016 to 27 Jan. 2016 (6 days)}
	\label{fig:case-nyc}
	\vspace{-10pt}
\end{figure}
\begin{figure}[t]
    \centering
    \includegraphics[width=1\linewidth, center]{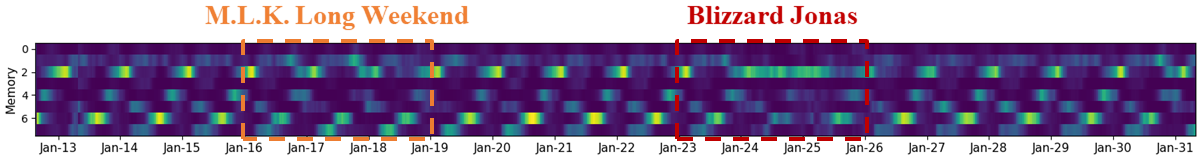}
    \vspace{-15pt}
    \caption{On-the-fly Attention Score ($\phi_j$) in MDFG at JONAS-NYC Test Set (from 12 Jan. to 31 Jan. 2016)}
    \label{fig:memo-att-nyc}
    \vspace{-15pt}
\end{figure}
\begin{figure}[t]
	\centering
	\begin{subfigure}{1\linewidth}
		\includegraphics[width=1\linewidth, center]{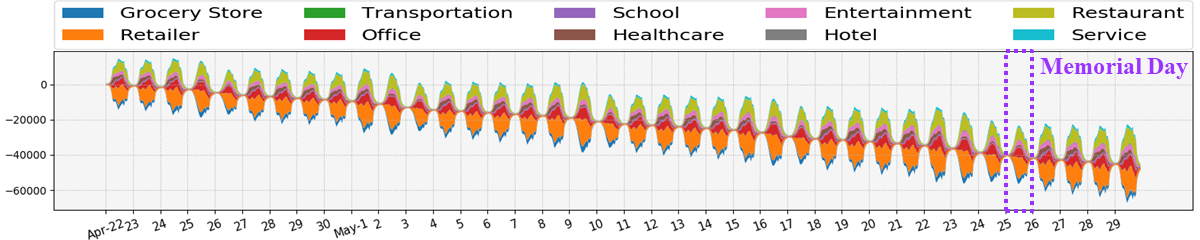}
		\vspace{-15pt}
		\caption{Ground Truth}
		\label{fig:case-us-true}
	\end{subfigure}
	\hspace{0.01cm}
	\begin{subfigure}{1\linewidth}
		\includegraphics[width=1\linewidth, center]{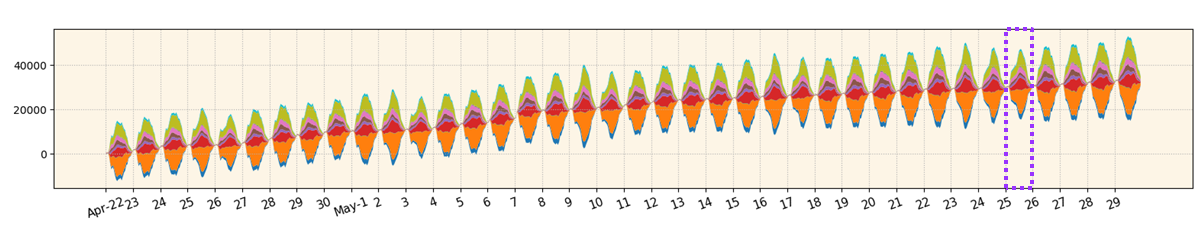}
		\vspace{-18pt}
		\caption{Prediction by GW-Net}
		\label{fig:case-us-sota}
	\end{subfigure}
	\hspace{0.01cm}
	\begin{subfigure}{1\linewidth}
		\includegraphics[width=1\linewidth, center]{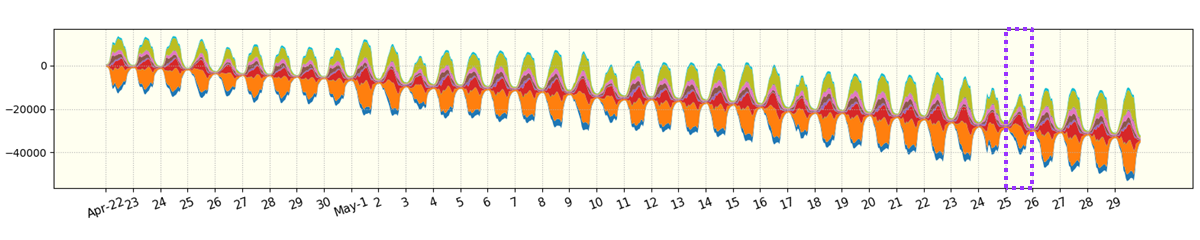}
		\vspace{-18pt}
		\caption{Prediction by EAST-Net}
		\label{fig:case-us-east}
	\end{subfigure}
	\vspace{-5pt}
	\caption{Stream Graphs of the Ground Truth and (4-hour ahead) Prediction Results of State-averaged POI Visits in 10 Categories from 22 Apr. 2020 to 29 May. 2020 (38 days)}
	\label{fig:case-us}
	\vspace{-10pt}
\end{figure}
\begin{figure}[t]
    \centering
    \includegraphics[width=1\linewidth, center]{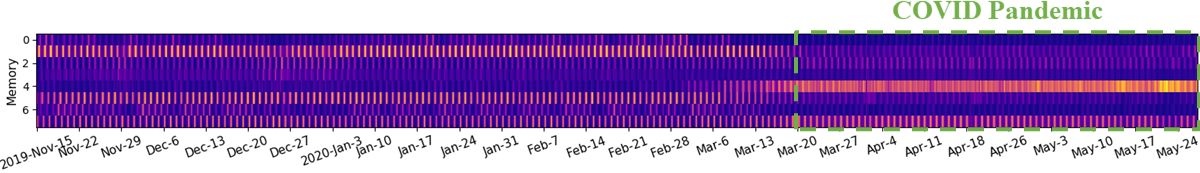}
    \vspace{-15pt}
    \caption{On-the-fly Attention Score ($\phi_j$) in MDFG at \\ COVID-US Set (from 15 Nov. 2019 to 25 May. 2020)}
    \label{fig:memo-att-us}
    \vspace{-15pt}
\end{figure}
\subsubsection{Qualitative Evaluation}
To understand how EAST-Net behaves in diverse scenarios of societal events, we conduct two case studies on JONAS-NYC and COVID-US.

In Figure \ref{fig:case-nyc}, a clear \textit{no-mobility} period is expected under the impact of the historic blizzard ``Jonas". GW-Net, a state-of-the-art model according to Table \ref{tab:baseline}, simply makes native forecasting (repeating the latest observation) during this anomalous period. In contrast, EAST-Net can quickly adapt to a \textit{declining-to-zero} tendency (although causing under-estimations afterwards). In addition, as illustrated in Figure \ref{fig:memo-att-nyc}, the composition of mobility prototypes in memory records for generating momentary filters is clearly differentiated between (1) normal workdays and weekend with a holiday; (2) a long weekend and the ``Jonas" period. These observations demonstrate the event-awareness and adaptivity of EAST-Net under a short-term event causing sudden volatility.

In Figure \ref{fig:case-us}, stream graphs \cite{byron2008stacked} for state averaged POI visits in ten categories during the first wave of COVID pandemic are presented. A stream graph is a variation of stacked area graph by positioning layers to minimize weighted wiggle (sum of the squared slopes). In our case, an overall negative ``tendency" is expected according to the ground truth. While an opposite positive ``tendency" is produced by GW-Net, EAST-Net can capture the overall shape correctly. In detail, on the Memorial Day (25 May. 2020), EAST-Net also better catch a obviously less volume of POI visits compared with GW-Net. In addition, according to Figure \ref{fig:memo-att-us}, EAST-Net actually became aware of the new mobility pattern as early as March, the very beginning of the epidemic in the US. These observations reconfirm the event-awareness and adaptivity of EAST-Net, particularly under a long-term event imposing lasting impact.

\section{Conclusion}
In this paper, we tackle the multimodal mobility nowcasting problem in response to various event scenarios. By designing a heterogeneous mobility information network for explicitly representing intermodality and a memory-augmented dynamic filter generator for producing sequence-specific parameters on-the-fly, we propose an event-aware spatio-temporal network (EAST-Net). Extensive experiments on four real-world datasets verify the event-awareness and adaptivity of EAST-Net, which is even applicable to unprecedented events. In the next step, we plan to improve the speed of adaptation and efficiency of EAST-Net.

\section{Acknowledgements}
This work was partially supported by JSPS KAKENHI (JP20K19859), JST Strategic International Collaborative Research Program (SICORP) (JPMJSC2002, JPMJSC2104), and Australian Research Council (ARC) Discovery Project (DP190101485). We are also appreciative of the open POI data (\textit{i.e.} Weekly Patterns) \href{https://www.safegraph.com/}{SafeGraph} has made.

\bibliography{./references.bib}

\begin{thebibliography}{39}
\providecommand{\natexlab}[1]{#1}

\bibitem[{Bahdanau, Cho, and Bengio(2014)}]{bahdanau2014neural}
Bahdanau, D.; Cho, K.; and Bengio, Y. 2014.
\newblock Neural Machine Translation by Jointly Learning to Align and
  Translate.
\newblock \emph{arXiv preprint arXiv:1409.0473}.

\bibitem[{Byron and Wattenberg(2008)}]{byron2008stacked}
Byron, L.; and Wattenberg, M. 2008.
\newblock Stacked Graphs--Geometry \& Aesthetics.
\newblock \emph{IEEE Transactions on Visualization and Computer Graphics},
  14(6): 1245--1252.

\bibitem[{Cao et~al.(2020)Cao, Wang, Duan, Zhang, Zhu, Huang, Tong, Xu, Bai,
  Tong, and Zhang}]{cao2020spectral}
Cao, D.; Wang, Y.; Duan, J.; Zhang, C.; Zhu, X.; Huang, C.; Tong, Y.; Xu, B.;
  Bai, J.; Tong, J.; and Zhang, Q. 2020.
\newblock Spectral Temporal Graph Neural Network for Multivariate Time-series
  Forecasting.
\newblock In \emph{Advances in Neural Information Processing Systems},
  17766--17778.

\bibitem[{Cirstea et~al.(2021)Cirstea, Kieu, Guo, Yang, and
  Pan}]{cirstea2021enhancenet}
Cirstea, R.-G.; Kieu, T.; Guo, C.; Yang, B.; and Pan, S.~J. 2021.
\newblock EnhanceNet: Plugin Neural Networks for Enhancing Correlated Time
  Series Forecasting.
\newblock In \emph{IEEE 37th International Conference on Data Engineering
  (ICDE)}, 1739--1750. IEEE.

\bibitem[{Fan et~al.(2019)Fan, Song, Jiang, Chen, and
  Shibasaki}]{fan2019decentralized}
Fan, Z.; Song, X.; Jiang, R.; Chen, Q.; and Shibasaki, R. 2019.
\newblock Decentralized Attention-based Personalized Human Mobility Prediction.
\newblock \emph{Proceedings of the ACM on Interactive, Mobile, Wearable and
  Ubiquitous Technologies}, 3(4): 1--26.

\bibitem[{Fan et~al.(2015)Fan, Song, Shibasaki, and
  Adachi}]{fan2015citymomentum}
Fan, Z.; Song, X.; Shibasaki, R.; and Adachi, R. 2015.
\newblock CityMomentum: An Online Approach for Crowd Behavior Prediction at a
  Citywide Level.
\newblock In \emph{Proceedings of the 2015 ACM International Joint Conference
  on Pervasive and Ubiquitous Computing}, 559--569.

\bibitem[{Feng et~al.(2018)Feng, Li, Zhang, Sun, Meng, Guo, and
  Jin}]{feng2018deepmove}
Feng, J.; Li, Y.; Zhang, C.; Sun, F.; Meng, F.; Guo, A.; and Jin, D. 2018.
\newblock DeepMove: Predicting Human Mobility with Attentional Recurrent
  Networks.
\newblock In \emph{Proceedings of the 2018 World Wide Web Conference},
  1459--1468.

\bibitem[{Ha, Dai, and Le(2016)}]{ha2016hypernetworks}
Ha, D.; Dai, A.; and Le, Q.~V. 2016.
\newblock Hypernetworks.
\newblock \emph{arXiv preprint arXiv:1609.09106}.

\bibitem[{Jia et~al.(2016)Jia, De~Brabandere, Tuytelaars, and
  Gool}]{jia2016dynamic}
Jia, X.; De~Brabandere, B.; Tuytelaars, T.; and Gool, L.~V. 2016.
\newblock Dynamic Filter Networks.
\newblock \emph{Advances in Neural Information Processing Systems}, 667--675.

\bibitem[{Jiang et~al.(2018)Jiang, Song, Fan, Xia, Chen, Miyazawa, and
  Shibasaki}]{jiang2018deepurbanmomentum}
Jiang, R.; Song, X.; Fan, Z.; Xia, T.; Chen, Q.; Miyazawa, S.; and Shibasaki,
  R. 2018.
\newblock DeepUrbanMomentum: An Online Deep-Learning System for Short-Term
  Urban Mobility Prediction.
\newblock In \emph{Thirty-Second AAAI Conference on Artificial Intelligence},
  784–791.

\bibitem[{Jiang et~al.(2019)Jiang, Song, Huang, Song, Xia, Cai, Wang, Kim, and
  Shibasaki}]{jiang2019deepurbanevent}
Jiang, R.; Song, X.; Huang, D.; Song, X.; Xia, T.; Cai, Z.; Wang, Z.; Kim,
  K.-S.; and Shibasaki, R. 2019.
\newblock DeepUrbanEvent: A System for Predicting Citywide Crowd Dynamics at
  Big Events.
\newblock In \emph{Proceedings of the 25th ACM SIGKDD International Conference
  on Knowledge Discovery \& Data Mining}, 2114--2122.

\bibitem[{Jiang et~al.(2021)Jiang, Wang, Cai, Yang, Fan, Xia, Matsubara,
  Mizuseki, Song, and Shibasaki}]{jiang2021countrywide}
Jiang, R.; Wang, Z.; Cai, Z.; Yang, C.; Fan, Z.; Xia, T.; Matsubara, G.;
  Mizuseki, H.; Song, X.; and Shibasaki, R. 2021.
\newblock Countrywide Origin-Destination Matrix Prediction and Its Application
  for COVID-19.
\newblock In \emph{Joint European Conference on Machine Learning and Knowledge
  Discovery in Databases}, 319--334. Springer.

\bibitem[{Konishi et~al.(2016)Konishi, Maruyama, Tsubouchi, and
  Shimosaka}]{konishi2016cityprophet}
Konishi, T.; Maruyama, M.; Tsubouchi, K.; and Shimosaka, M. 2016.
\newblock CityProphet: City-Scale Irregularity Prediction Using Transit App
  Logs.
\newblock In \emph{Proceedings of the 2016 ACM International Joint Conference
  on Pervasive and Ubiquitous Computing}, 752--757.

\bibitem[{Li et~al.(2019)Li, Jin, Xuan, Zhou, Chen, Wang, and
  Yan}]{li2019enhancing}
Li, S.; Jin, X.; Xuan, Y.; Zhou, X.; Chen, W.; Wang, Y.-X.; and Yan, X. 2019.
\newblock Enhancing the Locality and Breaking the Memory Bottleneck of
  Transformer on Time Series Forecasting.
\newblock \emph{Advances in Neural Information Processing Systems}, 32:
  5243--5253.

\bibitem[{Li et~al.(2018)Li, Yu, Shahabi, and Liu}]{li2018dcrnn_traffic}
Li, Y.; Yu, R.; Shahabi, C.; and Liu, Y. 2018.
\newblock Diffusion Convolutional Recurrent Neural Network: Data-Driven Traffic
  Forecasting.
\newblock In \emph{International Conference on Learning Representations (ICLR
  '18)}.

\bibitem[{Munkhdalai and Yu(2017)}]{munkhdalai2017meta}
Munkhdalai, T.; and Yu, H. 2017.
\newblock Meta Networks.
\newblock In \emph{International Conference on Machine Learning}, 2554--2563.
  PMLR.

\bibitem[{Pan et~al.(2019)Pan, Liang, Wang, Yu, Zheng, and
  Zhang}]{pan2019urban}
Pan, Z.; Liang, Y.; Wang, W.; Yu, Y.; Zheng, Y.; and Zhang, J. 2019.
\newblock Urban Traffic Prediction from Spatio-Temporal Data Using Deep Meta
  Learning.
\newblock In \emph{Proceedings of the 25th ACM SIGKDD International Conference
  on Knowledge Discovery \& Data Mining}, 1720--1730.

\bibitem[{Santoro et~al.(2016)Santoro, Bartunov, Botvinick, Wierstra, and
  Lillicrap}]{santoro2016meta}
Santoro, A.; Bartunov, S.; Botvinick, M.; Wierstra, D.; and Lillicrap, T. 2016.
\newblock Meta-Learning with Memory-Augmented Neural Networks.
\newblock In \emph{International Conference on Machine Learning}, 1842--1850.
  PMLR.

\bibitem[{Shi et~al.(2015)Shi, Chen, Wang, Yeung, Wong, and
  Woo}]{shi2015convolutional}
Shi, X.; Chen, Z.; Wang, H.; Yeung, D.-Y.; Wong, W.-K.; and Woo, W.-c. 2015.
\newblock Convolutional LSTM network: A machine learning approach for
  precipitation nowcasting.
\newblock In \emph{Advances in neural information processing systems},
  802--810.

\bibitem[{Song et~al.(2014)Song, Zhang, Sekimoto, and
  Shibasaki}]{song2014prediction}
Song, X.; Zhang, Q.; Sekimoto, Y.; and Shibasaki, R. 2014.
\newblock Prediction of Human Emergency Behavior and Their Mobility Following
  Large-Scale Disaster.
\newblock In \emph{Proceedings of the 20th ACM SIGKDD International Conference
  on Knowledge Discovery and Data Mining}, 5--14.

\bibitem[{Tang et~al.(2020)Tang, Yao, Sun, Aggarwal, Mitra, and
  Wang}]{tang2020joint}
Tang, X.; Yao, H.; Sun, Y.; Aggarwal, C.; Mitra, P.; and Wang, S. 2020.
\newblock Joint Modeling of Local and Global Temporal Dynamics for Multivariate
  Time Series Forecasting with Missing Values.
\newblock In \emph{Proceedings of the AAAI Conference on Artificial
  Intelligence}, 5956--5963.

\bibitem[{Vaswani et~al.(2017)Vaswani, Shazeer, Parmar, Uszkoreit, Jones,
  Gomez, Kaiser, and Polosukhin}]{vaswani2017attention}
Vaswani, A.; Shazeer, N.; Parmar, N.; Uszkoreit, J.; Jones, L.; Gomez, A.~N.;
  Kaiser, {\L}.; and Polosukhin, I. 2017.
\newblock Attention is All You Need.
\newblock In \emph{Advances in Neural Information Processing Systems},
  5998--6008.

\bibitem[{von Oswald et~al.(2020)von Oswald, Henning, Sacramento, and
  Grewe}]{Oswald2020Continual}
von Oswald, J.; Henning, C.; Sacramento, J.; and Grewe, B.~F. 2020.
\newblock Continual Learning with Hypernetworks.
\newblock In \emph{International Conference on Learning Representations (ICLR
  '20)}.

\bibitem[{Wang et~al.(2019)Wang, Yin, Chen, Wo, Xu, and Zheng}]{wang2019origin}
Wang, Y.; Yin, H.; Chen, H.; Wo, T.; Xu, J.; and Zheng, K. 2019.
\newblock Origin-Destination Matrix Prediction via Graph Convolution: A New
  Perspective of Passenger Demand Modeling.
\newblock In \emph{Proceedings of the 25th ACM SIGKDD International Conference
  on Knowledge Discovery \& Data Mining}, 1227--1235.

\bibitem[{Wang et~al.(2021)Wang, Xia, Jiang, Liu, Kim, Song, and
  Shibasaki}]{wang2021forecasting}
Wang, Z.; Xia, T.; Jiang, R.; Liu, X.; Kim, K.-S.; Song, X.; and Shibasaki, R.
  2021.
\newblock Forecasting Ambulance Demand with Profiled Human Mobility via
  Heterogeneous Multi-Graph Neural Networks.
\newblock In \emph{2021 IEEE 37th International Conference on Data Engineering
  (ICDE)}, 1751--1762. IEEE.

\bibitem[{Wu et~al.(2020)Wu, Pan, Long, Jiang, Chang, and
  Zhang}]{wu2020connecting}
Wu, Z.; Pan, S.; Long, G.; Jiang, J.; Chang, X.; and Zhang, C. 2020.
\newblock Connecting the Dots: Multivariate Time Series Forecasting with Graph
  Neural Networks.
\newblock In \emph{Proceedings of the 26th ACM SIGKDD International Conference
  on Knowledge Discovery \& Data Mining}, 753--763.

\bibitem[{Wu et~al.(2019)Wu, Pan, Long, Jiang, and Zhang}]{wu2019graph}
Wu, Z.; Pan, S.; Long, G.; Jiang, J.; and Zhang, C. 2019.
\newblock Graph WaveNet for Deep Spatial-Temporal Graph Modeling.
\newblock In \emph{Proceedings of the Twenty-Eighth International Joint
  Conference on Artificial Intelligence, {IJCAI-19}}, 1907--1913.

\bibitem[{Xie et~al.(2020)Xie, Guo, Chen, Xiao, Wang, and Zhao}]{xie2020deep}
Xie, Q.; Guo, T.; Chen, Y.; Xiao, Y.; Wang, X.; and Zhao, B.~Y. 2020.
\newblock Deep Graph Convolutional Networks for Incident-Driven Traffic Speed
  Prediction.
\newblock In \emph{Proceedings of the 29th ACM International Conference on
  Information \& Knowledge Management}, 1665--1674.

\bibitem[{Xue et~al.(2021)Xue, Salim, Ren, and Oliver}]{xue2021mobtcast}
Xue, H.; Salim, F.; Ren, Y.; and Oliver, N.~M. 2021.
\newblock MobTCast: Leveraging Auxiliary Trajectory Forecasting for Human
  Mobility Prediction.
\newblock In \emph{Thirty-Fifth Conference on Neural Information Processing
  Systems}.

\bibitem[{Yang et~al.(2019)Yang, Bender, Le, and Ngiam}]{yang2019condconv}
Yang, B.; Bender, G.; Le, Q.~V.; and Ngiam, J. 2019.
\newblock Condconv: Conditionally Parameterized Convolutions for Efficient
  Inference.
\newblock \emph{Advances in Neural Information Processing Systems}, 1307--1318.

\bibitem[{Yao et~al.(2019)Yao, Liu, Wei, Tang, and Li}]{yao2019learning}
Yao, H.; Liu, Y.; Wei, Y.; Tang, X.; and Li, Z. 2019.
\newblock Learning from Multiple Cities: A Meta-Learning Approach for
  Spatial-Temporal Prediction.
\newblock In \emph{The World Wide Web Conference}, 2181--2191.

\bibitem[{Yao et~al.(2018)Yao, Wu, Ke, Tang, Jia, Lu, Gong, Ye, and
  Li}]{yao2018deep}
Yao, H.; Wu, F.; Ke, J.; Tang, X.; Jia, Y.; Lu, S.; Gong, P.; Ye, J.; and Li,
  Z. 2018.
\newblock Deep Multi-View Spatial-Temporal Network for Taxi Demand Prediction.
\newblock In \emph{Proceedings of the AAAI Conference on Artificial
  Intelligence}, 2588–2595.

\bibitem[{Ye et~al.(2019)Ye, Sun, Du, Fu, Tong, and Xiong}]{ye2019co}
Ye, J.; Sun, L.; Du, B.; Fu, Y.; Tong, X.; and Xiong, H. 2019.
\newblock Co-Prediction of Multiple Transportation Demands based on Deep
  Spatio-Temporal Neural Network.
\newblock In \emph{Proceedings of the 25th ACM SIGKDD International Conference
  on Knowledge Discovery \& Data Mining}, 305--313.

\bibitem[{Zhang, Zheng, and Yu(2018)}]{zhang2018detecting}
Zhang, H.; Zheng, Y.; and Yu, Y. 2018.
\newblock Detecting urban anomalies using multiple spatio-temporal data
  sources.
\newblock \emph{Proceedings of the ACM on Interactive, Mobile, Wearable and
  Ubiquitous Technologies}, 2(1): 1--18.

\bibitem[{Zhang, Zheng, and Qi(2017)}]{zhang2017deep}
Zhang, J.; Zheng, Y.; and Qi, D. 2017.
\newblock Deep Spatio-Temporal Residual Networks for Citywide Crowd Flows
  Prediction.
\newblock In \emph{Thirty-first AAAI Conference on Artificial Intelligence},
  1655–1661.

\bibitem[{Zhang et~al.(2019)Zhang, Li, Shi, Li, Hui
  et~al.}]{zhang2019decomposition}
Zhang, M.; Li, T.; Shi, H.; Li, Y.; Hui, P.; et~al. 2019.
\newblock A Decomposition Approach for Urban Anomaly Detection across
  Spatiotemporal Data.
\newblock In \emph{Proceedings of the Twenty-Eighth International Joint
  Conference on Artificial Intelligence, {IJCAI-19}}, 6043--6049.

\bibitem[{Zheng et~al.(2020)Zheng, Fan, Wang, and Qi}]{zheng2020gman}
Zheng, C.; Fan, X.; Wang, C.; and Qi, J. 2020.
\newblock GMAN: A Graph Multi-Attention Network for Traffic Prediction.
\newblock In \emph{Proceedings of the AAAI Conference on Artificial
  Intelligence}, volume~34, 1234--1241.

\bibitem[{Zhou et~al.(2021)Zhou, Jampani, Pi, Liu, and
  Yang}]{zhou2021decoupled}
Zhou, J.; Jampani, V.; Pi, Z.; Liu, Q.; and Yang, M.-H. 2021.
\newblock Decoupled Dynamic Filter Networks.
\newblock In \emph{Proceedings of the IEEE/CVF Conference on Computer Vision
  and Pattern Recognition}, 6647--6656.

\bibitem[{Zonoozi et~al.(2018)Zonoozi, Kim, Li, and Cong}]{zonoozi2018periodic}
Zonoozi, A.; Kim, J.-j.; Li, X.-L.; and Cong, G. 2018.
\newblock Periodic-CRN: A Convolutional Recurrent Model for Crowd Density
  Prediction with Recurring Periodic Patterns.
\newblock In \emph{Proceedings of the Twenty-Eighth International Joint
  Conference on Artificial Intelligence, {IJCAI-18}}, 3732--3738.

\end{thebibliography}

\end{document}


\section{Additional Experiments}
\begin{figure*}[b]
	\centering
	\begin{subfigure}{1\linewidth}
		\includegraphics[width=1\linewidth, center]{./figures/fig-memo-att-1.PNG}
		\caption{on JONAS-NYC Test Set (from 12 Jan. to 31 Jan. 2016)}
		\label{fig:memo-att-nyc}
	\end{subfigure}
	\hspace{0.01cm}
	\begin{subfigure}{1\linewidth}
		\includegraphics[width=1\linewidth, center]{./figures/fig-memo-att-2.PNG}
		\caption{on COVID-US Set (from 15 Nov. 2019 to 25 May. 2020)}
		\label{fig:memo-att-us}
	\end{subfigure}
	\caption{Visualizations of Time-Variant Query Weights in Mobility Prototype Memory of EAST-Net During Two Studied Events}
	\label{fig:memo-att}
\end{figure*}

\subsection{Transferability}
To further evaluate the genericity of learnt representations in mobility prototype memory, we conduct two types of knowledge transfers, namely \textit{spatial transfer across regions} (under a same event), \textit{spatio-temporal transfer across events}. In Table \ref{tab:s-trans} and \ref{tab:st-trans}, \textbf{freeze} denotes the memory bank is loaded directly for testing; \textbf{retrain} denotes the memory values are loaded for initialization and then training the model again.

In Table \ref{tab:s-trans}, it can be observed that learnt knowledge from a source city needs to be further adapted to the target city, under the blizzard case. However, the COVID case is interestingly the opposite, despite that US and CHI have different spatio-temporal scales and measure different nature of mobility (transport mode \textit{v.s.} travel purpose). This phenomenon may be caused a shared low mobility status under the pandemic, so that the learnt patterns can be directly applied.
\begin{table}[h]
    \footnotesize
    \centering
    \caption{Spatial Knowledge Transfer: \{NYC, DC\} under JONAS and \{CHI, US\} under COVID}
    \label{tab:s-trans}
        \begin{tabular}{c||ccc} 
            \hline
            \textbf{Transfer} & \textit{RMSE} & \textit{MAE} & \textit{MAPE}  \\
            \hline
            DC \textbf{only} & 4.103 & 2.004 & 35.03\%    \\
            NYC$\rightarrow$DC \textbf{freeze} & 4.813 & 2.263 & 40.29\%      \\
            NYC$\rightarrow$DC \textbf{retrain} & \textbf{3.936} & \textbf{1.841} & \textbf{32.40\%}    \\
            \hline
            NYC \textbf{only} & \textbf{23.632} & \textbf{15.790} & \textbf{33.33\%}    \\
            DC$\rightarrow$NYC \textbf{freeze} & 36.372 & 25.537 & 48.96\%      \\
            DC$\rightarrow$NYC \textbf{retrain} & 24.062 & 16.150 & 34.59\%    \\
            \hline
            CHI \textbf{only} & 9.381 & 3.380 & \textbf{61.50\%}    \\
            US$\rightarrow$CHI \textbf{freeze} & \textbf{9.125} & \textbf{3.274} & 62.45\%       \\
            US$\rightarrow$CHI \textbf{retrain} & 11.687 & 4.691 & 78.98\%    \\
        \hline
    \end{tabular}
\end{table}
\begin{table}[h]
    \footnotesize
    \centering
    \caption{Spatio-Temporal Knowledge Transfer: JONAS-\{NYC, DC\} to COVID-CHI and JONAS-\{NYC, DC\} to COVID-US}
    \label{tab:st-trans}
    \addtolength{\tabcolsep}{-0.8ex}
        \begin{tabular}{c||ccc} 
            \hline
            \textbf{Transfer} & \textit{RMSE} & \textit{MAE} & \textit{MAPE}  \\
            \hline
            COVID-CHI \textbf{only} & 9.381 & 3.380 & 61.50\%    \\
            JONAS-NYC$\rightarrow$COVID-CHI \textbf{freeze} & 10.145 & 3.368 & 65.43\%      \\
            JONAS-NYC$\rightarrow$COVID-CHI \textbf{retrain} & \textbf{8.863} & \textbf{3.125} & 61.28\%    \\
            JONAS-DC$\rightarrow$COVID-CHI \textbf{freeze} & 9.229 & 3.560 & \textbf{60.51\%}   \\
            JONAS-DC$\rightarrow$COVID-CHI \textbf{retrain} & 10.922 & 3.795 & 79.75\%    \\
            \hline
            COVID-US \textbf{only} & \textbf{799.51} & \textbf{371.78} & \textbf{51.84\%}    \\
            JONAS-NYC$\rightarrow$COVID-US \textbf{freeze} & 3418.61 & 1476.85 & 461.48\%   \\
            JONAS-NYC$\rightarrow$COVID-US \textbf{retrain} & 1049.54 & 482.44 & 60.83\%     \\
            JONAS-DC$\rightarrow$COVID-US \textbf{freeze} & 2145.44 & 1046.92 & 319.29\%      \\
            JONAS-DC$\rightarrow$COVID-US \textbf{retrain} & 1003.05 & 456.91 & 60.77\%    \\
        \hline
    \end{tabular}
\end{table}

In Table \ref{tab:st-trans}, it can be observed that the nature of mobility matters when we are transferring over both spatial and temporal domains. Specifically, reusing historical lessons learnt from the blizzard Jonas can to some extent helps training or forecasting the multimodal transport demands under COVID pandemic, though the knowledge of DC is not as effective as the one of NYC for CHI. Not surprisingly, the knowledge learnt from different region and event for different nature of mobility is less mutually shareable, based on the results at COVID-US set.

\subsection{Interpretability}
To understand the inside mechanism of the proposed Memory-augmented Dynamic Filter Generator (MDFG), we chronologically visualize the time-variant query weights in mobility prototype memory during the two studied events. As illustrated in Figure \ref{fig:memo-att-nyc} and \ref{fig:memo-att-us}, some meaningful temporal patterns can also be interpreted and labeled. It is demonstrated that the memory learns to adapt to various scenarios, ranging from normal weekdays, weekends and holidays, to unprecedented events (\textit{i.e.} blizzard Jonas, COVID pandemic), for producing distinct model parameters on-the-fly.

\subsection{Efficiency}
In Table \ref{tab:efficiency}, we record and compare the efficiency of various models in terms of training and inference time on four datasets. Among all models, StemGNN and MTGNN are the two most efficient models as both are convolution-based temporal modeling. Despite the comprehensive view for intermodality, encoding the heterogeneous mobility information network (HMIN) is in fact less efficient than the conventional spatio-temporal view based on STG (in ST-Net). Compared with HMINet, EAST-Net has more intricate structures for memory query and filter generation, which consumes even more time. Overall, there is a tradeoff between the accuracy and efficiency in EAST-Net, which further motivates us to improve the model efficiency in the next step.
\begin{table*}[b]
    \footnotesize
    \centering
    \caption{Efficiency Comparison of Various Models in Training and Inference Time (\textit{s/epoch}) at JONAS-\{NYC, DC\}, COVID-\{CHI, US\}}
    \label{tab:efficiency}
    \addtolength{\tabcolsep}{-0.5ex}
        \begin{tabular}{c||cc|cc|cc|cc} 
            \hline
            \multirow{2}{*}{\textbf{Model}} & \multicolumn{2}{c|}{\textbf{JONAS-NYC}} & \multicolumn{2}{c|}{\textbf{JONAS-DC}} & \multicolumn{2}{c|}{\textbf{COVID-CHI}} & \multicolumn{2}{c}{\textbf{COVID-US}}  \\
            \cline{2-9}
            & Training & Inference & Training & Inference & Training & Inference & Training & Inference  \\
            \hline
            Transformer$^\dagger$ & 37.1 & 2.0 & 16.7 & 0.9 & 48.4 & 2.4 & 23.0 & 1.3      \\
            DCRNN & 37.4 & 2.1 & 19.1 & 1.1 & 47.6 & 2.6 & 31.6 & 1.6      \\
            GW-Net & 29.2 & 1.8 & 12.5 & 0.9 & 55.1 & 3.3 & 29.6 & 1.8      \\
            MTGNN & 17.7 & 1.2 & 8.3 & 0.7 & \textbf{25.0} & 1.9 & 18.4 & 1.3      \\
            StemGNN & \textbf{16.3} & \textbf{0.8} & \textbf{6.7} & \textbf{0.5} & 28.3 & \textbf{1.5} & \textbf{15.9} & \textbf{1.0}      \\
            \hline
            ST-Net & 28.5 & 1.7 & 15.0 & 0.9 & 36.2 & 2.0 & 24.6 & 1.5      \\
            HMINet & 51.9 & 2.7 & 24.5 & 1.4 & 67.4 & 3.9 & 47.2 & 2.5      \\
            \textbf{EAST-Net} & 60.2 & 3.5 & 30.5 & 1.9 & 72.1 & 4.6 & 55.3 & 3.1      \\
        \hline
    \end{tabular}
\end{table*}

\subsection{Data Preprocessing}
\begin{figure*}[t]
	\centering
    \includegraphics[width=0.9\linewidth, center]{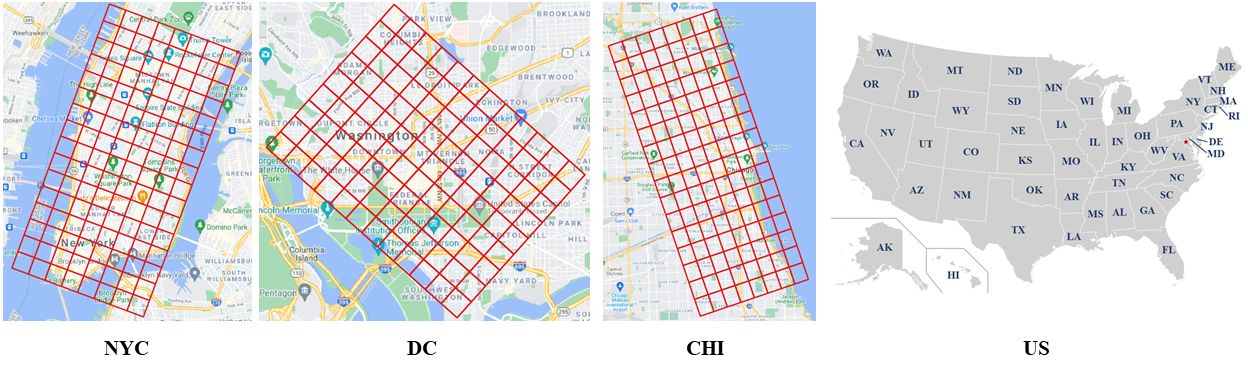}
	\caption{Spatial Settings of Four Experimental Datasets}
	\label{fig:spatial}
\end{figure*}

Figure \ref{fig:spatial} illustrates the spatial settings for discretization of four datasets respectively. Essentially, three city-level datasets (JONAS-NYC, JONAS-DC, COVID-CHI) are grid-based, while the country-level data (COVID-US) is graph-structured (each state is treated as a node and pairwise distance is considered as edges).

\section{Related Work}
In this section, we briefly review the existing researches in spatio-temporal forecasting especially for human mobility. Taking advantages of the recent advances in data acquisition and AI technologies, deep learning based approaches as a group has been applied on a variety of spatio-temporal prediction tasks, ranging from air quality \cite{yi2018deep}, incident \cite{huang2019mist, wu2020hierarchically, yuan2018hetero, wang2019cityguard, singh2018dynamic, zhao2017modeling}, to traffic volume \cite{yao2018deep, ye2019co, geng2019spatiotemporal} and traffic speed \cite{yu2018spatio, li2018dcrnn_traffic, guo2019attention, wu2019graph, diao2019dynamic, bai2019stg2seq, zhang2020spatio, cui2019traffic} and demonstrates its superiority over classical time series based models because of the complex nonlinear patterns of massive human behaviors. Specifically, for human mobility, there are two branches of studies, in which one branch focuses on individual/agent-based trajectory \cite{feng2018deepmove, fan2019decentralized}, and the other studies collective mobility behavior of crowds \cite{zhang2017deep, lin2019deepstn+, pan2019matrix}. Based on the data representation, the latter can be further broken down into two detailed categories: aggregated crowd flows \cite{zhang2017deep, lin2019deepstn+, pan2019matrix}, as well as origin-destination (OD) matrix \cite{wang2019origin, shi2020predicting}. In addition, there is line of research \cite{fan2015citymomentum, jiang2018deepurbanmomentum, jiang2019deepurbanevent} especially studies abnormal mobility prediction under events. Our work distinguishes itself by: (1) explicitly considering the multimodality of human mobility; (2) being capable to adapt to various scenarios for event-aware nowcasting.

\section{Definition Clarification}
Throughout the main content and appendix of this paper, we utilize the term "modal", "multimodal", "intermodal", "intermodality" particularly as derivatives of "mode" of mobility, or of transportation, rather than "modality of data" as in common usage.

\bibliography{./appendix-references.bib}